\documentclass[10pt,twocolumn,letterpaper]{article}
\usepackage[pagenumbers]{iccv} % To force page numbers, e.g. for an arXiv version
\usepackage{xcolor}
\definecolor{yzybest}{rgb}{0.96, 0.57, 0.58}
\definecolor{lightyellow}{rgb}{1.0, 1.0, 0.8}
\definecolor{warmyellow}{rgb}{1.0, 0.9, 0.6}
\definecolor{softyellow}{rgb}{1.0, 0.98, 0.85}
\usepackage{colortbl}
\usepackage{multirow}
\usepackage{pifont}
\newcommand{\paragrapht}[1]{\noindent\textbf{#1}}

\definecolor{iccvblue}{rgb}{0.21,0.49,0.74}
\usepackage[pagebackref,breaklinks,colorlinks,allcolors=iccvblue]{hyperref}

\def\confName{ICCV}
\def\confYear{2025}

\title{AM-Adapter: Appearance Matching Adapter \\ for Exemplar-based Semantic Image Synthesis in-the-Wild}

\author{
    Siyoon Jin$^{1}$  \qquad
    Jisu Nam$^{1}$ \qquad Jiyoung Kim$^{1}$ \qquad Dahyun Chung$^{2}$ \qquad \\
    Yeong-Seok Kim$^{3}$ \qquad Joonhyung Park$^{3}$ \qquad
    Heonjeong Chu$^{3}$ \qquad Seungryong Kim$^{1}$ \\ 
    $^{1}$KAIST \qquad $^{2}$Korea University \qquad $^{3}$Hyundai Mobis
}
\begin{document}
% \maketitle
\twocolumn[{%
\renewcommand\twocolumn[1][]{#1}%
\maketitle
\begin{center}
    \centering
    \captionsetup{type=figure}
    %\vspace{-20pt}
    \includegraphics[width=0.9\textwidth]{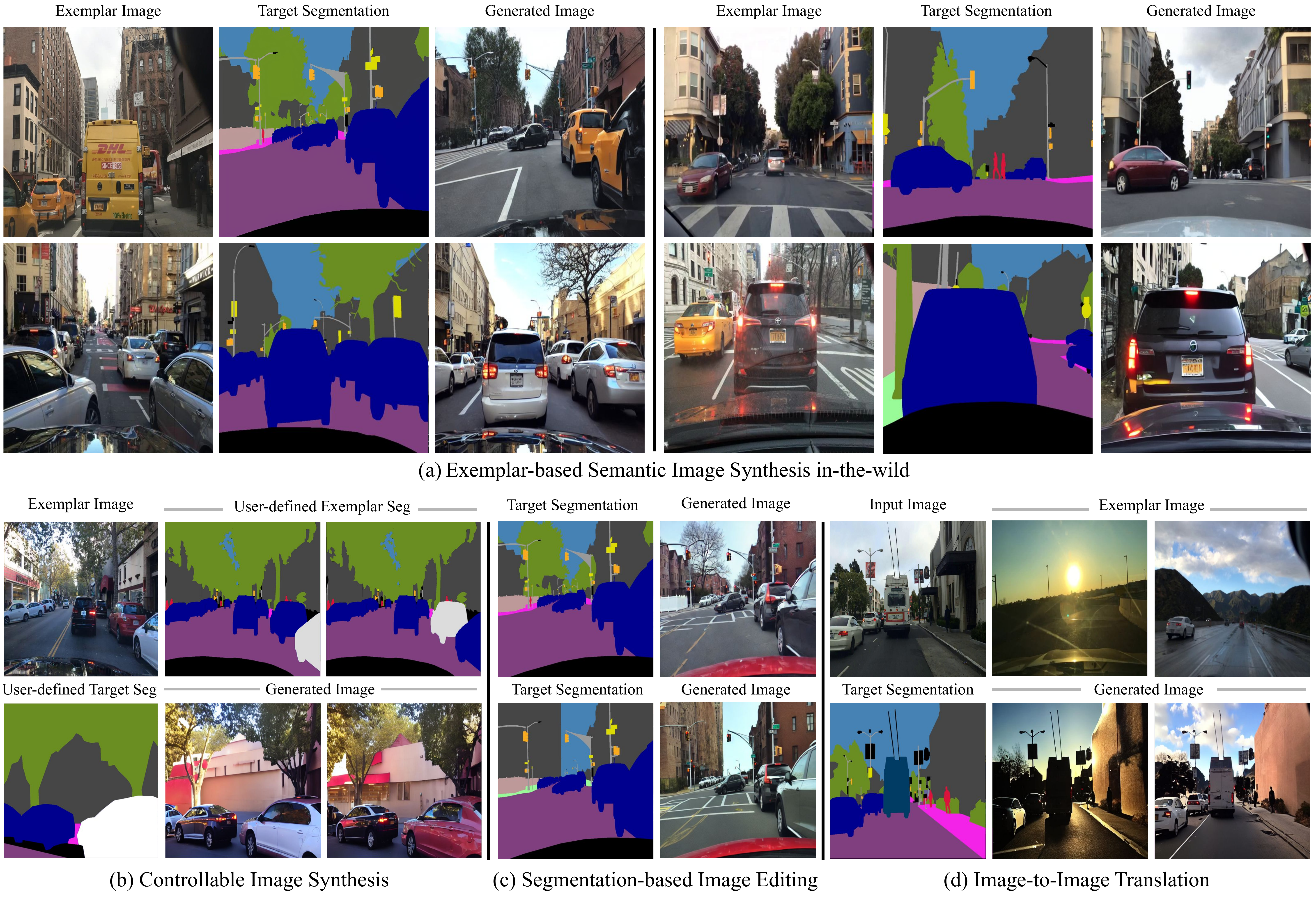}
    \vspace{-10pt}
    \captionof{figure}{\textbf{AM-Adapter enables Exemplar-based Semantic Image Synthesis in-the-Wild.} (a) Given an exemplar image and a target segmentation, \textbf{AM-Adapter} generates high-quality images that retain the local appearance of the exemplar and the accurate image structure defined by the segmentation map. We demonstrate the versatility of our method in various applications, including (b) controllable one-to-one appearance transfer with user-defined guidance, (c) image-to-image translation and (d) segmentation-based image editing.}
    \label{teaser}
% \vspace{2mm}
% \vspace{-5pt}
\end{center}%

}]
\begin{abstract}
Exemplar-based semantic image synthesis generates images aligned with semantic content while preserving the appearance of an exemplar. Conventional structure-guidance models like ControlNet, are limited as they rely solely on text prompts to control appearance and cannot utilize exemplar images as input. Recent tuning-free approaches address this by transferring local appearance via implicit cross-image matching in the augmented self-attention mechanism of pre-trained diffusion models. However, prior works are often restricted to single-object cases or foreground object appearance transfer, struggling with complex scenes involving multiple objects. To overcome this, we propose \textbf{AM-Adapter} (\textbf{A}ppearance \textbf{M}atching \textbf{Adapter}) to address exemplar-based semantic image synthesis in-the-wild, enabling multi-object appearance transfer from a single scene-level image. AM-Adapter automatically transfers local appearances from the scene-level input. AM-Adapter alternatively provides controllability to map user-defined object details to specific locations in the synthesized images. Our learnable framework enhances cross-image matching within augmented self-attention by integrating semantic information from segmentation maps. To disentangle generation and matching, we adopt stage-wise training. We first train the structure-guidance and generation networks, followed by training the matching adapter while keeping the others frozen. During inference, we introduce an automated exemplar retrieval method for selecting exemplar image-segmentation pairs efficiently. Despite utilizing minimal learnable parameters, AM-Adapter achieves state-of-the-art performance, excelling in both semantic alignment and local appearance fidelity. Extensive ablations validate our design choices. Code and weights will be released.: \url{https://cvlab-kaist.github.io/AM-Adapter/}
\end{abstract}
\vspace{-15pt}

% many-to-many appearance transfer across diverse categories.     
\section{Introduction}
\label{sec:introduction}

\begin{figure*}
    \centering
    %\vspace{-20pt}
    \includegraphics[width=\textwidth]{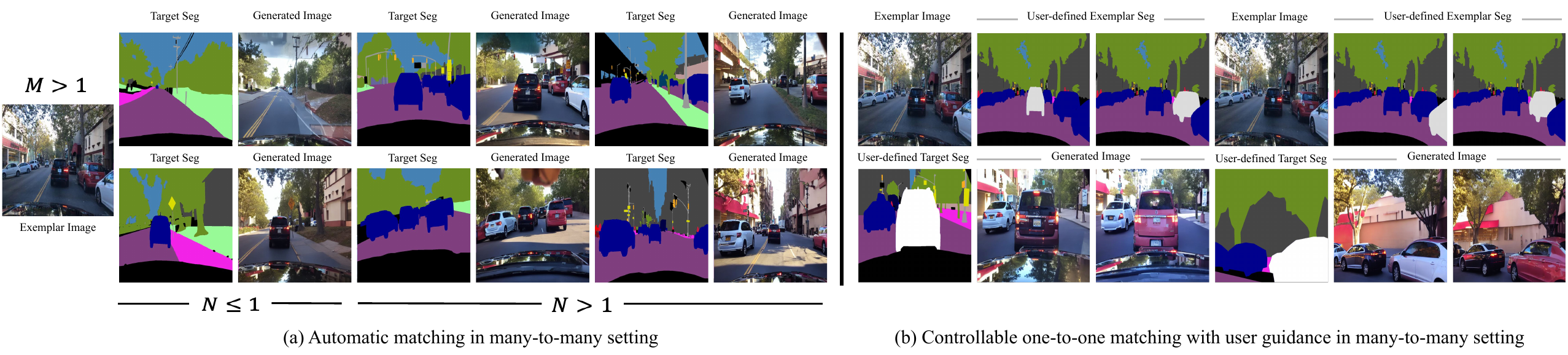} 
    \vspace{-15pt}
    \caption{\textbf{Controllability of our \textbf{AM-Adapter}:} The $M$-to-$N$ setting refers to a multiple-object many-to-many transfer, where $M$ and $N$ denote the number of instances of a specific category in the exemplar and target, respectively. By default, as shown in (a), AM-Adapter automatically matches appearance based on structural similarity. In addition, as shown in (b), it allows user-defined guidance to precisely transfer specific objects (e.g., red car) to target instances  (White-masked objects indicate user-specified matches). 
} 
    % if a specific object (e.g., red car) needs to be transferred to a particular target instance, user-defined guidance allows precise control over the matching process. White-masked objects are user-specified ones.}

    \vspace{-10pt}
    
    \label{qual:matching_case}
\end{figure*}

Exemplar-based semantic image synthesis~\cite{ye2023ip, alaluf2024cross, mo2024freecontrol, lin2024ctrl} aims to generate an image aligned with given semantic content (e.g., segmentation maps) while faithfully reflecting user-defined visual attributes (e.g., styles, lighting, or object details). This task is critical for various downstream applications, including autonomous driving~\cite{eskandar2023pragmaticsemanticimagesynthesis}, medical imaging~\cite{zhuang2023semanticimagesynthesisabdominal}, and virtual/augmented reality (AR/VR)~\cite{habtegebrial2020generativeviewsynthesissingleview}.

Recent advances in structure-guidance methods~\cite{ zhao2024uni, peng2024controlnext, li2025controlnet} such as ControlNet allow fine-grained spatial control but rely on text prompts (e.g., `realistic' or `nighttime') for appearance. As shown in Figure~\ref{qual:comp}, they often fail to transfer exemplar-specific details, as text alone is insufficient to convey fine-grained visual information.

% by integrating an additional encoder into pre-trained text-to-image diffusion models, while adjusting visual appearance based on text prompts (e.g., `realistic' or `nighttime'). However, despite their promising structural control, as shown in Figure~\ref{qual:intro}(a), they often fail to capture user-defined visual details, as text descriptions alone are insufficient to convey fine-grained visual information.

To address this, several methods have utilized exemplars for visual conditioning~\cite{ye2023ip, alaluf2024cross, lin2024ctrl}, leveraging pre-trained image encoders such as CLIP~\cite{radford2021learningtransferablevisualmodels}. This is achieved by mapping the exemplar's features from the encoder onto the pre-trained diffusion model using linear projection heads~\cite{ye2023ip, li2024photomaker, gal2023encoder, xiao2024fastcomposer, peng2024portraitbooth}. However, as shown in Figure~\ref{qual:comp}, these models still struggle to accurately transfer local appearances from exemplars, as global mappings via linear layers are insufficient for embedding complex visual features. As a result, they are often limited to global appearance transfer while failing to capture fine-grained, object-specific details.% into generation models.

To resolve these, some studies~\cite{alaluf2024cross, lin2024ctrl, cao2023masactrl, nam2024dreammatcher} have explored local appearance transfer by hand-crafted attention control in self-attention modules of pre-trained diffusion models. Prior works~\cite{cao2023masactrl, nam2024dreammatcher} achieve this through augmented self-attention, providing the key and value matrices of the exemplar to the self-attention that synthesize an image aligned with the given semantic content. This can be seen as finding correspondences between synthesized and exemplar images via implicit query-key matching, then warping appearance using the corresponding value matrices. While effective in constrained settings, they struggle in complex, real-world scenes with multiple objects and geometric variations. As shown in Figure~\ref{qual:comp}, implicit matching often misaligns categories (e.g., cars to buildings, trees to roads), causing artifacts and distortions. Recent methods introduce additional constraints~\cite{nam2024dreammatcher, alaluf2024cross, lin2024ctrl}, yet remain limited to single or foreground object cases, making them ineffective for multi-object many-to-many scenarios.

% These methods  transfer the local appearances of exemplars to the synthesized image in a constrained setting. However, as presented in Figure~\ref{qual:intro}(c), they rely on implicit matching within the self-attention mechanism, leading to inaccurate results under large geometric deformations in content-rich, complex scenarios (e.g., driving scenes). This further causes inconsistent correspondences across different categories (e.g., cars matched to buildings, trees matched to roads), ultimately producing blurry or distorted outputs.

% Recent approaches~\cite{nam2024dreammatcher,alaluf2024cross, lin2024ctrl} address this by applying additional constraints to enhance the attention signal, but they remain limited to foreground objects, hindering their applicability in complex scenarios.

% Recent methods introduce additional constraints~\cite{nam2024dreammatcher, alaluf2024cross, lin2024ctrl}, they remain limited to single-object cases or object-centric one-to-many transfers, making them ineffective for multi-object many-to-many scenarios.
 
The most straightforward solution is to fine-tune the augmented self-attention module to improve matching. However, it is challenging as the module needs to learn both matching and generation simultaneously, leading to an unstable training signal and overfitting, as discussed in Figure \textcolor{NavyBlue}{12}(d). While recent studies~\cite{hu2024animate} attempted to fine-tune such model, they are typically limited to specific domains, such as foreground-centric human videos~\cite{hu2024animateanyoneconsistentcontrollable}.
% In contrast, we introduce Exemplar-based Semantic Synthesis in-the-Wild, the task to enable many-to-many appearance transfer across multiple object categories. To this end, we propose \textbf{AM-Adapter}, a learnable module 

% To this end, we present a novel method that combines the spatial control capabilities of structure-guidance model~\cite{peng2024controlnext} with effective local appearance transfer via learnable appearance matching control. We introduce the \textbf{Appearance Matching Adapter (AM-Adapter)}, which enhances implicit matching within the augmented self-attention of pre-trained diffusion models by integrating semantic information from segmentation maps (Figure~\ref{qual:intro}(d)). Built on a frozen pre-trained diffusion model and ControlNeXt~\cite{peng2024controlnext}, the adapter focuses solely on enhancing matching, effectively disentangling generation from matching to maximize performance.

Importantly, the aforementioned methods fail to capture object-specific styles, as they either apply style globally, ignoring local details, or focus on single/foreground object. We define this task as \textit{exemplar-based semantic image synthesis in-the-wild}, extending appearance transfer to complex, scene-level multi-object transfer. 
% that focus on single-object or foreground often struggle with multi-object appearance transfer in complex scenes. Existing approaches often fail to capture individual object styles (e.g, transferring car \textit{color} to car) within a single image, as they transfer style globally, which differ object-by-object.  We define this task as \textit{Exemplar-based Semantic Image Synthesis In-the-Wild}, extending it to scene-level. 
To solve this, we propose \textbf{AM-Adapter}, which enhances implicit matching in augmented self-attention of pre-trained diffusion by integrating semantics from segmentation maps. This enables both global and local appearance transfer through semantic correspondence and structural similarity, while preserving scene composition and ensuring natural object-wise appearance transfer. 
% Built on a frozen pre-trained diffusion model and ControlNeXt~\cite{peng2024controlnext}, our approach disentangles matching from generation, improving stability and fidelity.
% We build our adapter on top of the pre-trained diffusion model and ControlNeXt~\cite{peng2024controlnext}, freezing all its parameters to fully leverage its generative capability and structural consistency, while the adapter focuses solely on enhancing matching. This design effectively disentangles generation from matching, maximizing matching performance.

Our architecture consists of two branches: \textit{Appearance Net}, which extracts local appearance from an exemplar, and \textit{Structure Net}, which synthesizes a target image from random noise, aligned with a target segmentation map. To harness the semantic relationship between exemplar and target images for improved matching, we first construct a categorical matching cost between the exemplar and target segmentation maps, which is then concatenated with the implicit matching cost in self-attention. To find locally consistent matches within the 4D space between the exemplar and target, we refine the combined cost using a 4D cost aggregation module. The final output is added residually to the original implicit matching cost for stable training.

To further disentangle generation and matching, we adopt a stage-wise training approach. First, we pre-train the diffusion model and ControlNeXt~\cite{peng2024controlnext} on the target domain, and then train the matching adapter while keeping the diffusion model and ControlNeXt frozen. This approach is highly efficient, as it only requires updating the matching adapter's parameters, preserving the image quality and structural consistency of the pre-trained models.

During inference, previous studies~\cite{lin2024ctrl, alaluf2024cross} manually select exemplar images, which may be labor-intensive in a specific setting. To address this, we propose an automatic exemplar retrieval method that selects exemplar image-segmentation pairs from a large pool to maximize matchable regions with the synthesized target image.

Despite the limited number of learnable parameters, our method achieves state-of-the-art performance in complex segmentation-to-image tasks, such as driving scenes~\cite{yu2020bdd100k, cordts2016cityscapes}. Extensive ablation studies validate the effectiveness of each component, and we demonstrate the versatility of our model across various downstream applications, as in Figure~\ref{teaser}. Our method eliminates manual style assignment by automatically matching appearances based on scene structure (Figure~\ref{qual:matching_case}(a)). Additionally, it also supports user-defined one-to-one matching, allowing precise control over specific object correspondences (Figure~\ref{qual:matching_case}(b)).
%The code and checkpoint will be made publicly available.

\begin{figure*}
    \centering
    %\vspace{-20pt}
    \includegraphics[width=\textwidth]{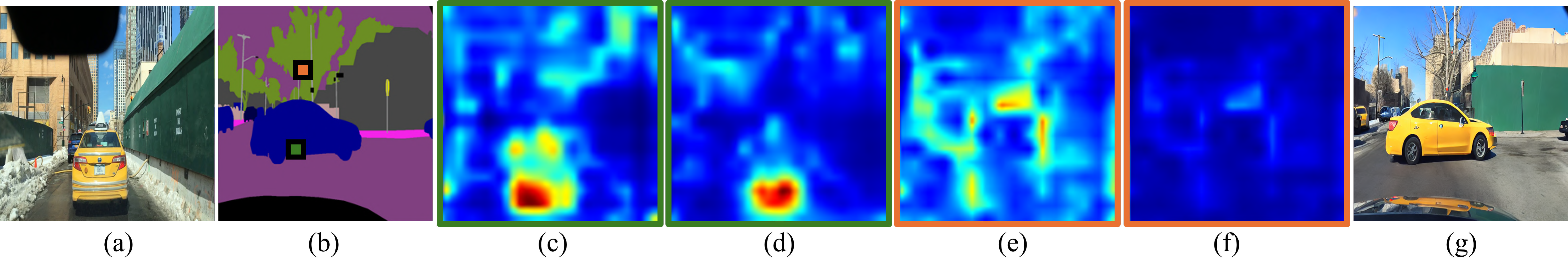} 
    \vspace{-15pt}
    \caption{\textbf{Attention Visualization:} (a) Exemplar image, (b) target segmentation, and (g) generated image. \textcolor{green}{Green} and \textcolor{orange}{orange} markers in (b) indicate query points. (c) and (d) show the augmented self-attention map $Q_t^Y (K_t^X)^T$ from the {green} marker, before and after applying AM-Adapter, respectively. (e) and (f) show the augmented self-attention map $Q_t^Y (K_t^X)^T$ from the {orange} marker, before and after applying AM-Adapter, respectively. The {green} marker is within `car' (present in the reference), while the {orange} marker is within `tree' (absent in the reference). \textbf{AM-Adapter} refines mismatches in (c) and (e), demonstrating its effectiveness in (d) and (f).}
    %(a) and (b) represent the exemplar image containing the desired appearance and the target segmentation encapsulating the structure, respectively. The green marker and orange marker in (b) present the query point in the target segmentation. (c), (d), (e), and (f) represent augmented self-attention $Q_t^Y (K_t^X)^T$ at the query point. Specifically, (c) and (d) show the attention maps before and after applying adapter for the green marker in (b), while (e) and (f) do so for the orange marker. The green marker is within the query object 'car', present in the reference, while the orange marker is within the 'tree', absent in the reference. Applying the AM-Adapter refines random mismatches to irrelevant categories, resulting in improved alignment and demonstrating its effectiveness.} 

    %The third and fourth columns show the augmented self-attention, $Q_t^Y (K_t^X)^T$, before and after applying the adapter, respectively. In the first row, the query object, "car", is present in the reference. After applying the adapter, mismatches in categories are refined, resulting in a smoother attention map. In the second row,  the query object, "tree", is absent from the reference. Prior to applying the adapter, random matches to irrelevant categories are prominent. However, after applying the adapter, these mismatches are fully refined, demonstrating improved alignment. These results demonstrate the effectiveness and validity of the AM-Adapter.}
    \vspace{-15pt}
    
    \label{qual:attn}
\end{figure*}

\section{Related Work}
\label{sec:relwork}

\paragrapht{Semantic Image Synthesis.}
Semantic Image Synthesis~\cite{li2025controlnet, peng2024controlnext, zhao2024uni, zhang2023adding, park2019semantic, wang2022semantic, mou2024t2i, li2023gligen, lv2024placeadaptivelayoutsemanticfusion, xue2023freestylelayouttoimagesynthesis} aims to generate images that align with given semantic structures. Several models~\cite{li2025controlnet, peng2024controlnext, mou2024t2i, li2023gligen} provide structural guidance using a learnable spatial encoder integrated into the pre-trained diffusion model and control global appearance through text prompts. However, these models often fail to capture user-defined appearance details, as they do not incorporate exemplar images, and text descriptions alone are insufficient to depict fine-grained details in the images.

\paragrapht{Exemplar-based Semantic Image Synthesis.}
Several studies~\cite{ye2023ip, li2024photomaker, gal2023encoder, xiao2024fastcomposer, peng2024portraitbooth, hu2024instructimagenimagegenerationmultimodal} have leveraged pre-trained image encoders, such as CLIP~\cite{radford2021learningtransferablevisualmodels}, to extract image features from an exemplar and project them into compact embeddings, often through linear projection layers. While these methods achieve promising results in specific domains, they fail to preserve local appearance in content-rich scenes, since they focus on global appearance.

Recent studies address this with hand-crafted attention control, injecting exemplar keys and values into the self-attention module of the target. This transfers the exemplar value, which includes appearance information, into the target structure based on the matching cost between the target query and exemplar key. Several methods~\cite{nam2024dreammatcher, lin2024ctrl, alaluf2024cross} enhance this matching cost by using intermediate diffusion features~\cite{nam2024dreammatcher}, spatially-aware appearance transfer~\cite{lin2024ctrl}, or attention map contrasting~\cite{alaluf2024cross}. While these approaches improve implicit matching, performance remains limited by the inherent matching accuracy of self-attention, which is still constrained to foreground-centric objects and struggles with precise matches in content-rich images.

\section{Method}
\label{sec:method}

\subsection{Preliminaries}
Diffusion models adopt a UNet architecture~\cite{rombach2022high} with an encoder and decoder, where each block includes self- and cross-attention layers. Cross-attention is guided by a text condition, while self-attention captures intra-feature relationships. At each timestep \( t \), self-attention projects features into queries \( Q_t \), keys \( K_t \), and values \( V_t  \) of shape \( \mathbb{R}^{h \times w \times d} \), where $h$, $w$, and $d$ indicate height, width, and channels, respectively. The attention score is formulated as:
% , and the attention score is calculated by matrix multiplication between $Q_t$ and $K_t$.
% In the self-attention layer, the intermediate feature at each timestep \( t \) is projected into queries \( Q_t \in \mathbb{R}^{h \times w \times d} \), keys \( K_t \in \mathbb{R}^{h \times w \times d} \), and values \( V_t \in \mathbb{R}^{h \times w \times d} \), where $h$, $w$, and $d$ indicate height, width, and channels, respectively, and the attention score is calculated by matrix multiplication between $Q_t$ and $K_t$. 
% Then, $V_t$ is warped to the queries based on the calculated attention score, formulated as:
%\begin{equation}
%    \mathtt{S\texttt{-}Att}(Q_t, K_t, V_t) =  \mathtt{Softmax}(A_t) V_t,\quad  A_t = \frac{Q_t K_t^T}{\sqrt{d}},
%\end{equation}
\begin{equation}
    \mathtt{S\texttt{-}Att}(Q_t, K_t, V_t) =  \mathtt{Softmax}(A_t) V_t, \quad  A_t = \frac{Q_t K_t^T}{\sqrt{d}}.
\label{eq:self_att}
\end{equation}
where $\mathtt{S\texttt{-}Att} (\cdot)$ denotes the self-attention operation. $A_t \in \mathbb{R}^{h \times w \times h \times w}$ are attention scores. $\mathtt{Softmax}(\cdot)$ normalizes the attention scores across the keys.

\subsection{Problem Statement and Overview}
Given an exemplar image $I^{{X}}\in \mathbb{R}^{H \times W \times 3}$ and a target segmentation map $S^{{Y}}\in \mathbb{R}^{H \times W \times 1}$, where $H$ and $W$ denote height and width, our goal is to generate a target image $I^{{Y}}$ that retains the appearance of $I^{{X}}$, while maintaining the structure of $S^{{Y}}$. Prior works such as ControlNet~\cite{zhang2023adding} or related variants~\cite{zhao2024uni, peng2024controlnext, li2025controlnet} provide structural guidance (e.g., segmentation maps) and adapt appearance via text control~\cite{zhang2023adding} or an additional image encoder~\cite{ye2023ip,gal2023encoder}. However, they fail to retain local details from $I^{{X}}$ as they either do not take the exemplar image as input~\cite{zhang2023adding, peng2024controlnext} or compress local features into a global representation~\cite{ye2023ip,gal2023encoder}.

\begin{figure*}
    \centering
    %\vspace{-20pt}
    \includegraphics[width=0.95\linewidth]{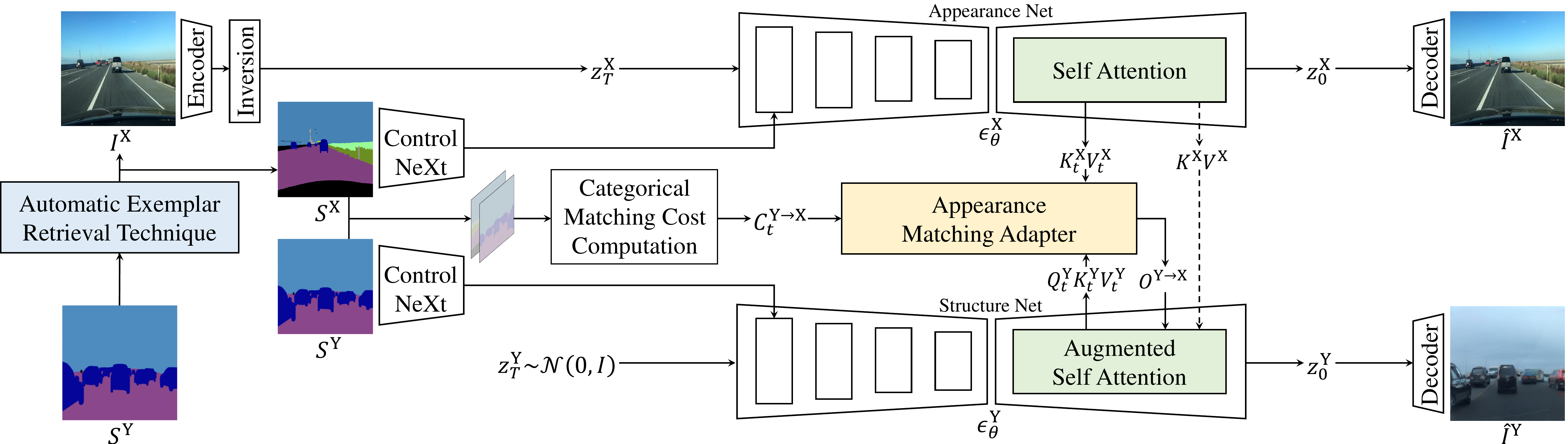} 
    \vspace{-10pt}
    \caption{\textbf{Overall Architecture.}}
    \vspace{-15pt}    
    \label{qual:architecture}
\end{figure*}

To address this, we aim to enhance local appearance from $I^{{X}}$ in the target image $I^{{Y}}$, while retaining the target structure from $S^{{Y}}$. We propose a dual-branch architecture: Appearance Net $\epsilon_\theta^X(\cdot)$ and Structure Net $\epsilon_\theta^Y(\cdot)$. Specifically, $I^{X}$ is encoded via VAE~\cite{pu2016variational} into $z_0^X$ then inverted into $z_T^X$ by DDIM inversion~\cite{mokady2023null}. $z_T^X$ is subsequently reconstructed to $z_0^X$ by the Appearance Net, guided by $S^{X}$. The Structure Net synthesizes $\hat{I}^{Y}$ from random Gaussian noise $z_T^Y$ to $z_0^Y$, conditioned on  $S^{Y}$. Both $z_0^X$ and $z_0^Y$ are decoded as $\hat{I}^X$ and $\hat{I}^Y$, respectively, via VAE~\cite{pu2016variational}. At each time step, exemplar keys and values, $K^X_t$ and $V^X_t$, from the Appearance Net are injected into the augmented self-attention of the Structure Net. Our AM-Adapter enhances the matching between $Q^Y_t$ from Structure Net and $K^X_t$ from Appearance Net, reflecting the semantic relationship between $S^{Y}$ and $S^{X}$. The architecture is illustrated in Figure~\ref{qual:architecture}.

\begin{figure}[t]
    \centering
    \includegraphics[width=0.9\linewidth]{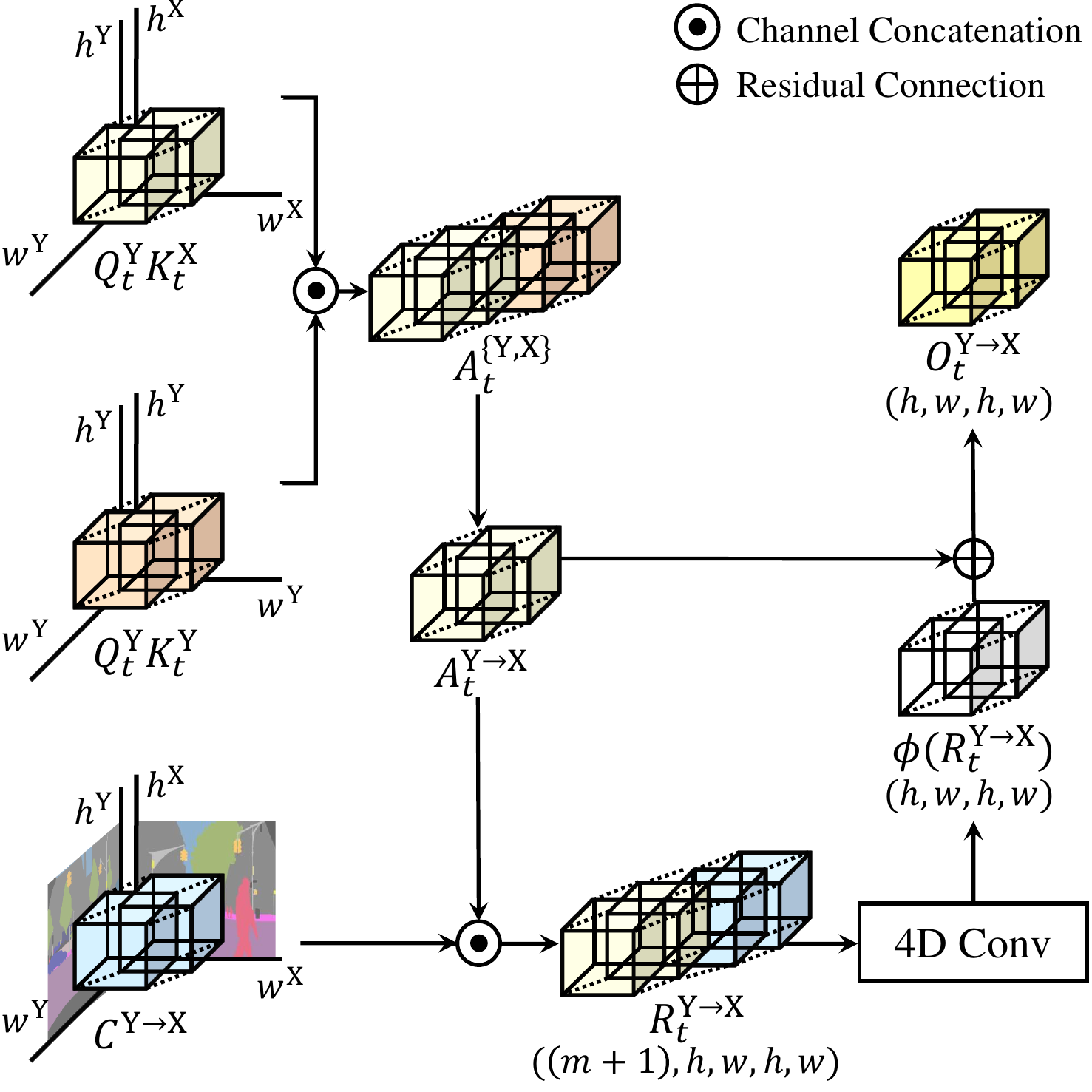} 
    %\vspace{-10pt}
\caption{\textbf{Architecture of Appearance Matching Adapter.}}
    \vspace{-10pt}
    \label{adapter}
\end{figure}

\subsection{Baseline and Analysis}
\label{sec:baseline}
The most straightforward method to transfer appearance from Appearance Net to the Structure Net, as in previous works~\cite{lin2024ctrl, nam2024dreammatcher, cao2023masactrl}, could be to naively concatenate the exemplar keys $K_t^X$ and values $V_t^X$ with the corresponding target keys $K_t^Y$ and values $V_t^Y$, which can formulate as: 
% \begin{equation}
%     \label{equ:sa-aug}
%     \mathtt{S\texttt{-}Att}(Q_t^{{Y}}, K_t^\mathrm{C}, V_t^\mathrm{C}) = \mathtt{Softmax}(A^\mathrm{C}_t) \cdot V_t^\mathrm{C},
% \end{equation}
% \begin{equation}
%  A^\mathrm{C}_t = \frac{Q_t^{Y} (K_t^\mathrm{C})^T}{\sqrt{d}},
% \end{equation}
% \begin{equation}
%  K_t^\mathrm{C} = \mathtt{Concat}(K_t^Y, K_t^X),
% \end{equation}
% \begin{equation}
%  V_t^\mathrm{C} = \mathtt{Concat}(V_t^Y, V_t^X),
% \end{equation}
% where $\mathtt{Concat}(\cdot)$ indicates the concatenation operation. 
\begin{equation}
    \label{equ:sa-aug}
    \mathtt{S\texttt{-}Att}(Q_t^{{Y}}, K_t^{\{Y,X\}}, V_t^{\{Y,X\}}) = \mathtt{Softmax}(A^{\{Y,X\}}_t) V_t^{\{Y,X\}},
\end{equation}
\begin{equation}
 A^{\{Y,X\}}_t = {Q_t^{Y} (K_t^{\{Y,X\}})^T}/{\sqrt{d}},
\end{equation}
\begin{equation}
 K_t^{\{Y,X\}} = \mathtt{Concat}(K_t^Y, K_t^X),
\end{equation}
\begin{equation}
 V_t^{\{Y,X\}} = \mathtt{Concat}(V_t^Y, V_t^X),
\end{equation}
where $\mathtt{Concat}(\cdot)$ indicates the concatenation. 
We divide $A^{\{Y,X\}}_t$ into $A^{Y \rightarrow Y}_t$ and $A^{Y \rightarrow X}_t$, where $A^{Y \rightarrow Y}_t$ captures intra-feature relationships in the target image, and $A^{Y \rightarrow X}_t$ represents the matching cost between target queries $Q^Y_t$ and exemplar keys $K^X_t$. Here, $Q^Y_t$ attends to $K^X_t$ to selectively integrate exemplar values $V_t^X$. As noted in~\cite{alaluf2024cross}, this process transfers local appearance details from $V_t^X$ to the target image based on the semantic matching  between $Q_t^{Y}$ and $K_t^X$.
% We divide $A^{\{Y,X\}}_t$ into $A^{Y \rightarrow Y}_t$ and $A^{Y \rightarrow X}_t$, where $A^{Y \rightarrow Y}_t$ is the matching cost between $Q^Y_t$ and $K^Y_t$, which learns the intra-relationships of the features in the synthesized image, while $A^{Y \rightarrow X}_t$ is the matching cost between the target queries $Q^Y_t$ and the exemplar keys $K^X_t$, where the target queries $Q^Y_t$ attend to the exemplar keys $K^X_t$ and selectively integrate the corresponding exemplar values $V_t^X$. As discussed in~\cite{alaluf2024cross}, this process can be analyzed by transferring local appearance details in $V_t^X$ into the target image based on the semantic matching  between $Q_t^{Y}$ and $K_t^X$.

However, as shown in Figure \textcolor{NavyBlue}{12}(b), this approach often produces suboptimal results, as it relies on implicit matching within the self-attention trained for generation, leading to inaccurate matches, particularly in complex domains. In particular, this results in mismatches across distinct categories (e.g., cars being matched to trees). 

One solution to address this is to directly train the augmented self-attention module to enhance the implicit matching in the pre-trained diffusion model. However, as shown in Figure \textcolor{NavyBlue}{12}(c), this approach is challenging, as the model needs to learn both generation and matching at the same time, resulting in unstable training and overfitting.

\subsection{Appearance Matching Adapter}
\label{sec:adapter_detail}
To address this, we propose a learnable appearance matching adapter, AM-Adapter that enhances the implicit matching $A^{Y \rightarrow X}_t$ with semantic awareness in a data-driven manner. The detailed architecture is illustrated in Figure~\ref{adapter}.

\paragrapht{Categorical Matching Cost.} To obtain improved matching, it is crucial to be aware of the categorical relationship between the exemplar and target images, as pixels in the same category serve as strong initial points for establishing accurate correspondence. To this end, we first compute a categorical matching cost by calculating pixel-wise correspondences between the segmentation maps of the exemplar and target images, denoted by $S^{{X}}$ and $S^{{Y}}$, respectively.

\begin{figure}[t!]
    \centering
    % \vspace{-20pt}
    \includegraphics[width=1\linewidth]{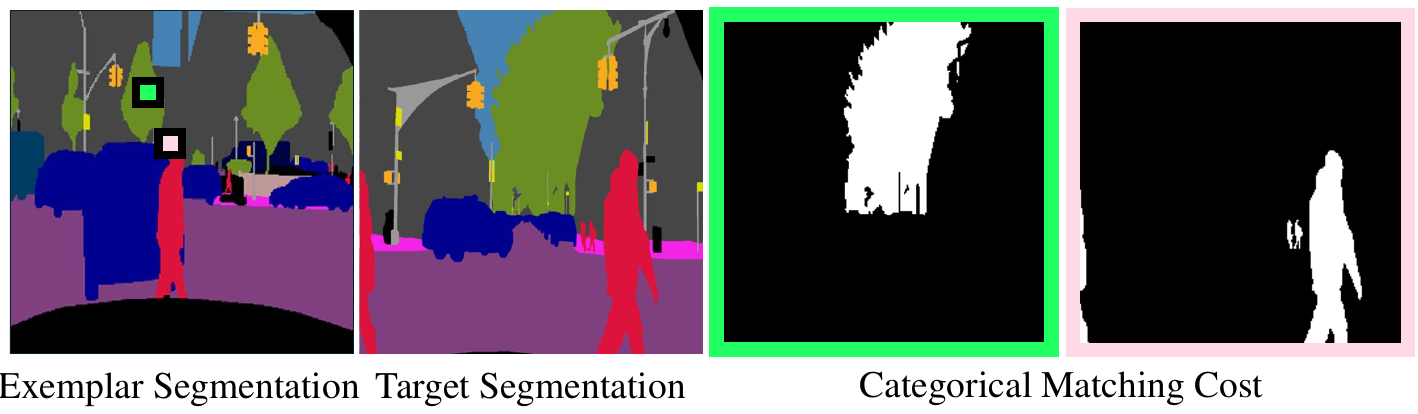} 
    \vspace{-15pt}
    \caption{\textbf{Visualization of Categorical Matching Cost.} \textcolor{green}{Green} and \textcolor{pink}{pink} markers denote query points in the exemplar segmentation map. Our categorical matching cost effectively identifies binary categorical matches between exemplar and target segmentation maps (best viewed in color).}
    \vspace{-15pt}
    
    \label{qual:categorical}
\end{figure}

Given segmentation maps $S^X$ and $S^Y$ with semantic labels, we build the binary categorical matching cost $C^{Y\rightarrow X}$ by identifying pixel positions in $S^Y$ with the categorical index of each pixel in $S^X$. This process is formulated as:
\begin{equation}
    C^{Y\rightarrow X}(i, j, k, l) = 
    \begin{cases}
        1, & \text{if $S^Y(i,j) = S^X(k,l)$},\\
        0, & \text{otherwise},
    \end{cases}
\end{equation}
where $i,j \in [0, H) \times [0, W)$ and $k,l \in [0, H) \times [0, W)$. % and $\mathbbm{1}$ denotes the binary indicator.
Figure~\ref{qual:categorical} visualizes $C^{Y\rightarrow X}$.

\paragrapht{Attention Aggregation Module.} The most straightforward approach to apply the categorical matching cost $C^{Y\rightarrow X}$ is to directly multiply implicit matching cost $A_t^{Y\rightarrow X}$ by $C^{Y\rightarrow X}$ during the inference. However, as shown in Figure \textcolor{NavyBlue}{12}(d), while this method can filter out mismatches across categories, it often fails to find the correct correspondences within the same category. This is because $C^{Y\rightarrow X}$ is a binary map that does not directly aligned with the attention scores in $A_t^{Y\rightarrow X}$ within the self-attention module.

% To address this, inspired by prior works~\cite{hong2022neuralmatchingfieldsimplicit, min2021hypercorrelationsqueezefewshotsegmentation}, we propose a lightweight 4D convolution network to refine the implicit matching cost $A_t^{Y\rightarrow X}$ with the categorical matching cost $C^{Y\rightarrow X}$ in a data-driven manner, enabling more precise pixel-wise correspondence between exemplar and target. Note that, unlike 2D convolution networks, which process each image independently and do not capture local relationships between the exemplar and target, 4D convolution achieves reliable matching by identifying locally consistent regions across both images in 4D space.
To address this, inspired by~\cite{hong2022neuralmatchingfieldsimplicit, min2021hypercorrelationsqueezefewshotsegmentation}, we propose a lightweight 4D convolution network to refine the implicit matching cost $A_t^{Y\rightarrow X}$ using the categorical matching cost $C^{Y\rightarrow X}$ in a data driven manner. This enables more precise pixel-wise correspondence between exemplar and target. Unlike 2D convolutions, which process image independently, 4D convolution ensure reliable matching by capturing locally consistent regions across images in 4D space.

Specifically, at each time step $t$, we concatenate the implicit matching cost $A_t^{Y\rightarrow X}$ and the categorical matching cost $C^{Y\rightarrow X}$ along the channel dimension. Here, to allow each self-attention head to learn distinct relationships, we retain the number of multiple heads, denoted as $m$, and thus $A_t^{Y\rightarrow X} \in \mathbb{R}^{h \times w \times h \times w \times (m+1)}$. This results in $R_t^{Y \rightarrow X} \in \mathbb{R}^{h \times w \times h \times w \times (m+1)}$, where $h$, $w$ and $m+1$ represent the height, width and concatenated channel dimension, respectively. $C^{Y\rightarrow X}$ is downsampled to match the dimensions of $A_t^{Y\rightarrow X}$. The process is formulated as: 
\begin{equation}
    R_t^{Y \rightarrow X} = \mathtt{Concat}(A_t^{Y\rightarrow X}, \mathtt{Downsample}(C^{Y\rightarrow X})),
\end{equation}
where $\mathtt{Concat}$ denotes concatenation operation along the channel dimension, while $\mathtt{Downsample}$ indicates the downsampling.
To enhance the matching cost with semantic awareness, the 4D convolution network aggregates the concatenated matching cost $R_t^{Y \rightarrow X}$ and outputs the refined matching cost. To stabilize the learning process, the refined cost is added to the original implicit matching cost $A_t^{Y\rightarrow X}$ through a residual connection, resulting in the final output $O_t^{Y \rightarrow X} \in \mathbb{R}^{h \times w \times h \times w}$. The process is formulated as:
\begin{equation}
O_t^{Y \rightarrow X} = \phi(R_t^{Y \rightarrow X}) + A_t^{Y\rightarrow X},
\end{equation}
Here, $\phi(\cdot)$ denotes our learnable 4D convolution network. In Figure~\ref{qual:attn}, we compare the attention maps from $A_t^{Y\rightarrow X}$ and $O_t^{Y \rightarrow X}$, highlighting the impact of our 4D cost aggregation module. 
Finally, in the Structure Net, $A^{\{Y,X\}}_t$ at each time step $t$ is enhanced by combining $O^{Y \rightarrow Y}_t$ and $A^{Y \rightarrow X}_t$.

In addition, our framework allows user-defined guidance to precisely transfer specific objects to target instances. For this, we manipulate the categorical cost to restrict target's selected one to attend only to exemplar's selected one, ensuring precise focus on the designated region.

\begin{figure*}
    \centering
    \vspace{-20pt}
    \includegraphics[width=\linewidth]{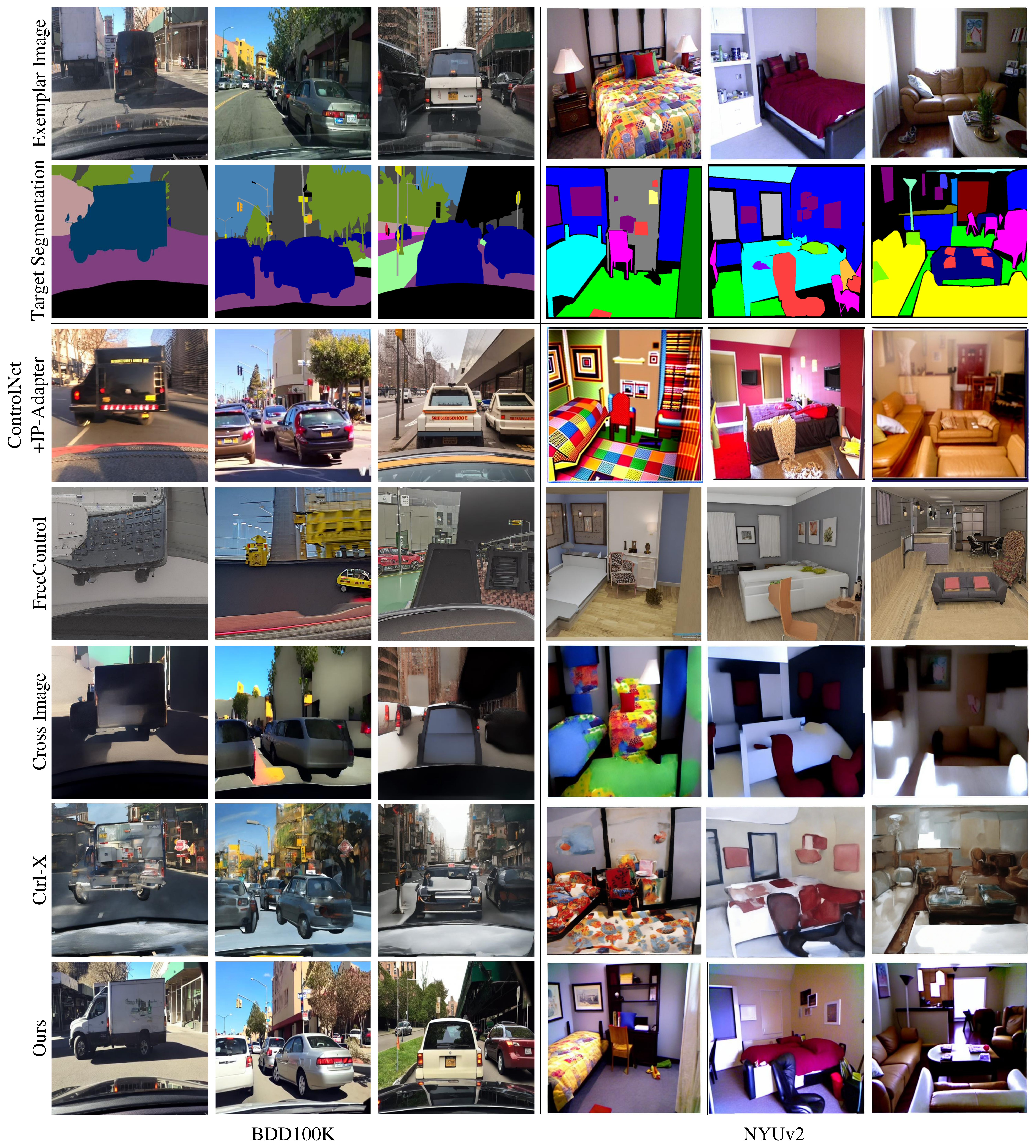} 
    \vspace{-15pt}
    % \caption{\textbf{Qualitative Comparison on BDD100K~\cite{yu2020bdd100k} and Cityscapes~\cite{cordts2016cityscapes}:} Compared to previous methods~\cite{zhang2023adding, ye2023ip, mo2024freecontrol, alaluf2024cross, lin2024ctrl} that fail to transfer local appearance or find accurate matching, AM-Adapter effectively achieves precise matching in content-rich, complex scenes. }
    \caption{\textbf{Qualitative Comparison on BDD100K~\cite{yu2020bdd100k} and NYUv2~\cite{Silberman:ECCV12}:} Compared to previous methods\cite{zhang2023adding, ye2023ip, mo2024freecontrol, alaluf2024cross, lin2024ctrl} that fail to transfer local appearance or find accurate matching, AM-Adapter effectively achieves precise matching in content-rich, complex scenes.  Furthermore, our approach generalizes well across diverse domains (driving and indoor scene), ensuring robust appearance transfer.}
    \vspace{-15pt}
    
    \label{qual:comp}
\end{figure*}

\subsection{Training}
\label{sec:training}
\paragrapht{Exemplar-Target Pair.}
To train for local appearance transfer from the exemplar to the target, it is essential for both images to share similar local features. We achieve this via data augmentation to create exemplar-target pairs. Specifically, we apply different augmentations, such as random cropping~\cite{Takahashi_2020} and flipping~\cite{zhou2022flipdaeffectiverobustdata}, to an anchor image to generate exemplar-target pairs $(I^X, I^Y)$. The same augmentations are applied to the segmentation maps $(S^X, S^Y)$. Using these pairs, we train the AM-Adapter to guide the pre-trained diffusion model in generating target images that reflect the local appearance of the exemplar. 
%This is formulated as predicting the added noise in the diffusion forward process, $\epsilon \sim \mathcal{N}(0,1)$, as in~\cite{ho2020denoising, song2020denoising}:
%\begin{equation} 
%\mathcal{L} := \mathbb{E}_{\epsilon,z^Y_t,t,S^Y, \phi(R_t^{Y \rightarrow X})} \left[ \left\lVert \epsilon - \epsilon_{\theta} (z^Y_t, t, S^Y, \phi(R_t^{Y \rightarrow X}) \right\rVert^2_2 \right],
%\end{equation}
%where $z^Y_t$ is the latent variable of $I^Y$ at time step $t$, and $\epsilon_{\theta}(\cdot)$ represents the pre-trained T2I diffusion model. 

% random crop / flip

%During inference, previous studies~\cite{} manually select an exemplar image, which is labor-intensive. To address this, we propose an automatic exemplar retrieval method that selects exemplar image-segmentation pairs from a large pool to maximize matchable regions with the synthesized target image.

% \input{author-kit-CVPR2025-v3.1-latex-/tables/1_quan_comp}

% \input{author-kit-CVPR2025-v3.1-latex-/figures/7_qual_comp}
\begin{figure*}[htb!]
    \centering
    \vspace{-20pt}
    \includegraphics[width=\textwidth]{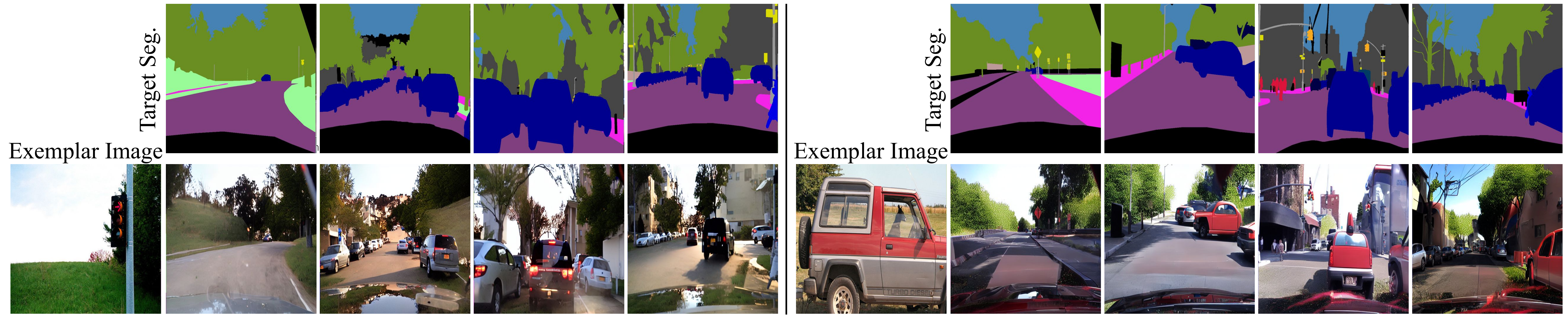} 
    \vspace{-10pt}
    % \caption{\textbf{Robustness to Exemplar.} If the exemplar lacks structural similarity, the model ignores it. AM-Adapter performs robust even when the exemplar and target are structurally non-homogeneous.}
    \caption{\textbf{Robustness to Exemplar.} If the exemplar lacks structural similarity to the target segmentation maps, AM-Adapter robustly ignores it and remains effective on non-homogeneous pairs.}
    %\vspace{-10pt}
    \label{qual:exemplar}
\end{figure*}

\begin{table*}[t]
\small
% \vspace{-20pt}
\centering
\resizebox{0.9\textwidth}{!}{
    \begin{tabular}{l|c|ccccc}
    \toprule
    Methods & Training & Self-sim. $(\downarrow)$ & $I_\mathrm{DINO} (\uparrow)$ & DINO [cls] loss $(\downarrow)$ & $I_\mathrm{CLIP} (\uparrow)$ & FID $(\downarrow)$ \\
    % \multirow{2}{*}{Methods} & \multirow{2}{*}{Training} & \multicolumn{3}{c|}{BDD100K~\cite{yu2020bdd100k}} & \multicolumn{3}{c}{Cityscapes~\cite{cordts2016cityscapes}}  \\
    % \cmidrule(lr){3-5}\cmidrule(lr){6-8}& & Self-sim. $(\downarrow)$ & $I_{\mathrm{CLIP}}$ $(\uparrow)$  &FID $(\downarrow)$ & Self-sim. $(\downarrow)$ &  $I_{\mathrm{CLIP}}$ $(\uparrow)$ & FID $(\downarrow)$  \\
    % % \cmidrule(lr){3-5}\cmidrule(lr){6-8}
    \midrule
    ControlNet~\cite{zhang2023adding} 
    & \checkmark & 0.052 & 0.654 & 0.095 & 0.638 & 95.10   \\
    ControlNeXt~\cite{peng2024controlnext}  
    & \checkmark & \cellcolor{warmyellow}{0.047} & 0.695 & 0.078 & 0.731 & 94.48  \\
    \midrule
    FreeControl~\cite{mo2024freecontrol}  & \ding{55} & 0.062 & 0.479 & 0.100 & 0.575 & 179.00   \\
    Cross-Image Attention~\cite{alaluf2024cross}  & \ding{55} & 0.177 & 0.495 & 0.137 & 0.643  & 233.76   \\
    Ctrl-X~\cite{lin2024ctrl}  & \ding{55} & 0.051 & \cellcolor{warmyellow}{0.743} & \cellcolor{warmyellow}{0.063} & 0.700  & 107.06 \\
    \midrule

    ControlNet~\cite{zhang2023adding} + IP-Adapter~\cite{ye2023ip} & \checkmark & 0.049 & 0.732 & 0.072 & \cellcolor{warmyellow}{0.805} & \cellcolor{warmyellow}{84.82}  \\
    %ControlNet~\cite{zhang2023adding} + DreamMatcher~\cite{nam2024dreammatcher} & \checkmark & \cellcolor{warmyellow}{0.049} & 0.759 & \cellcolor{warmyellow}{0.805} & \cellcolor{warmyellow}{84.82} & \cellcolor{warmyellow}{0.049} & 0.783 & \cellcolor{warmyellow}{0.823} & 115.98  \\
    
    \midrule
    \textbf{AM-Adapter (Ours)} & \checkmark &  \cellcolor{yzybest}{0.041} & \cellcolor{yzybest}{0.765} & \cellcolor{yzybest}{0.056} & \cellcolor{yzybest}{0.819} & \cellcolor{yzybest}{75.89}  \\
    \bottomrule
    \end{tabular}
}
% \vspace{-10pt}
\caption{\textbf{Quantitative Comparison.}}
\vspace{-15pt}
\label{tab:main}
\end{table*}

\paragrapht{Stage-wise Training.} The most direct approach to train our model is joint training of ControlNeXt~\cite{peng2024controlnext}, the diffusion model, and AM-Adapter.
However, this entangles structural guidance (ControlNeXt), generation (diffusion) and matching (AM-Adapter), leading to instability, overfitting and degraded performance.
% However, this poses challenges in disentangling structural guidance from ControlNeXt, the generative capability of the diffusion model, and matching process from the adapter, leading to unstable training and potential overfitting, which degrades structural consistency, image quality, or matching performance. 
To mitigate this, we adopt stage-wise training: first, we train ControlNeXt on image-segmentation pairs for fine-grained structural control. Next, to generate realistic samples in the desired domain, we train the diffusion model on the domain with ControlNeXt frozen. Finally, we train AM-Adapter on frozen Appearance Net and Structure Net, using the same pre-trained diffusion model and ControlNeXt. The stage-wise approach effectively separates structural guidance, generation, and matching, achieving high performance in each component.

\subsection{Inference}
\label{sec:inference}
\paragrapht{Retrieval-based Inference.} During inference, previous tuning-free methods~\cite{alaluf2024cross, lin2024ctrl} require manual exemplar selection. However, identifying suitable exemplars for generating content-rich scene-level images, is labor-intensive. To address this, we introduce an automatic exemplar retrieval technique. 
%Our matching adapter accurately identifies correspondences, particularly when the exemplar’s structure closely aligns with the target structure. 
A basic approach selects the closest structural match from a pool of segmentation maps and uses the corresponding RGB image as the exemplar. However, we find that segmentation-based retrieval often misses finer details, resulting in suboptimal outcomes. To improve precision, we retrieve exemplar based on structural similarity in RGB space. Specifically, we generate a random RGB image from the target segmentation map using Structure Net, preserving detailed semantic information. We then convert it to grayscale, comparing it with grayscale images in the pool to find the most structurally similar exemplar. This process yields the top-1 RGB exemplar image for each target segmentation map. Even if the retrieved exemplar lacks structural similarity, the model ignores it; greater similarity enhances appearance transfer. Figure~\ref{qual:exemplar} demonstrates that AM-Adapter remains effective even for non-homogeneous image pairs. %The effectiveness of retrieval-based inference is analyzed in Tab~\ref{tab:abl} and Fig.~\ref{qual:abl}. 
More details and ablation are in Appendix \textcolor{NavyBlue}{C.2}.

% \begin{table*}[t]
% % \small
% \centering
% \begin{tabular}{l|c|ccc|ccc|}
% \toprule
% \multirow{4}{*}{Methods} & {Training} & \multicolumn{3}{c|}{BDD100K~\cite{}} & \multicolumn{3}{c|}{CityScapes~\cite{}}  \\
% \cmidrule(lr){3-5}\cmidrule(lr){6-8}& & Self-sim $\downarrow$ & $I_{\mathrm{CLIP}}$ $\downarrow$ & FID $\downarrow$ & Self-sim $\downarrow$ & $I_{\mathrm{CLIP}}$ $\downarrow$ & FID $\downarrow$  \\
% % \cmidrule(lr){3-5}\cmidrule(lr){6-8}
% \midrule
% ControlNet~\cite{} + IPAdapter~\cite{} & \checkmark & -- & -- & -- & -- & -- & -- \\
% FreeControl~\cite{}  & \ding{55} & -- & -- & -- & -- & -- & -- \\
% Cross-Image Attention~\cite{}  & \ding{55} & -- & -- & -- & -- & -- & --  \\
% Ctrl-X~\cite{}  & \ding{55} & -- & -- & -- & -- & -- & -- \\
% % \midrule

% % \midrule
% \bluerow \textbf{Ours} & -- &  -- & -- & -- & -- & -- \\
% \bottomrule
% \end{tabular}
% \caption{\textbf{Quantitative comparisons.}}
% \label{tab:main}
% \end{table*}

\begin{table*}[t]
\small
\vspace{-20pt}
\centering
\resizebox{0.9\textwidth}{!}{
    \begin{tabular}{l|l|ccccc}
    \toprule
    & & Self-sim. $(\downarrow)$ & $I_\mathrm{DINO} (\uparrow)$ & DINO[cls] loss $(\downarrow)$ &$I_{\mathrm{CLIP}}$ $(\uparrow)$  &FID $(\downarrow)$   \\
    % \cmidrule(lr){3-5}\cmidrule(lr){6-8}
    % 
    \midrule
    &  Exemplar image & 0.052 & 1.000 & 0.000 & 1.000 & 44.23 \\
    \midrule 
    (I) & Pre-training~\cite{peng2024controlnext} & 0.047 & 0.695 & 0.063 & 0.731 & 94.48  \\
    (II) & (I) + MasaCtrl~\cite{cao2023masactrl}
     & 0.047 & \cellcolor{yzybest}{0.848} & \cellcolor{yzybest}{0.039} & \cellcolor{yzybest}{0.872} &  79.45\\
    (III) & (I) + DreamMatcher~\cite{nam2024dreammatcher}
    &  0.046 & 0.693 & 0.070 & 0.724  &79.34 \\
    
    \midrule
    (IV) & (I) + Augmented Self-Attention & 0.043 & 0.751 & 0.057 & 0.815  & 78.13  \\ 
    (V) & (IV) + w/ Fine-tuning & \cellcolor{softyellow}{0.043} & 0.747 & 0.058 & 0.808 & \cellcolor{softyellow}{77.75}  \\ 
    (VI) & (IV) + Categorical Matching Cost & \cellcolor{warmyellow}{0.043} & \cellcolor{softyellow}{0.751} & \cellcolor{softyellow}{0.056} & \cellcolor{softyellow}{0.817} & \cellcolor{warmyellow}{76.76}  \\
    \midrule 
    % (VII) & AM-Adapter (Ours) w/o matching guidance. & -- & -- & -- & -- & -- & -- \\
    (VII) &  \textbf{AM-Adapter (Ours)} &   \cellcolor{yzybest}{0.041}  & \cellcolor{warmyellow}{0.765} & \cellcolor{warmyellow}{0.056} & \cellcolor{warmyellow}{0.819} & \cellcolor{yzybest}{75.89}  \\
    \bottomrule
    \end{tabular}
}
% \vspace{-15pt}
\caption{\textbf{Ablation on Individual Components.} Although the table displays scores rounded to three decimal places, we determine the rankings based on precision to the fourth decimal place. Additional results and analysis on ablations are in Appendix.}
\vspace{-15pt}
\label{tab:abl_quan}
\end{table*}

% \begin{table*}[t]
% \small
% \centering
% \resizebox{\textwidth}{!}{
%     \begin{tabular}{l|c|cccc|cccc}
%     \toprule
%     \multirow{2}{*}{Methods} & \multirow{2}{*}{Training} & \multicolumn{4}{c|}{BDD100K~\cite{yu2020bdd100k}} & \multicolumn{4}{c|}{CityScapes~\cite{cordts2016cityscapes}}  \\
%     \cmidrule(lr){3-6}\cmidrule(lr){7-10}& & Self-sim $\downarrow$ & $I_{\mathrm{DINO}}$ $\uparrow$ & $I_{\mathrm{CLIP}}$ $\uparrow$  &CLIP-FID $\downarrow$ & Self-sim $\downarrow$ & $I_{\mathrm{DINO}}$ $\uparrow$ & $I_{\mathrm{CLIP}}$ $\uparrow$ & CLIP-FID $\downarrow$  \\
%     % \cmidrule(lr){3-5}\cmidrule(lr){6-8}
%     \midrule

% \input{ICCV2025-Author-Kit-Feb/figures/6_qual}
% \input{ICCV2025-Author-Kit-Feb/tables/1_main_quan}
\paragrapht{Matching Cost Guidance.} To further enhance the matching between the exemplar and target, we propose a matching cost guidance technique during inference. Specifically, we apply classifier-free guidance~\cite{ho2022classifierfreediffusionguidance} on the refined matching cost $O_t^{Y \rightarrow X}$, which is formulated as:
\begin{equation}
    \tilde{\epsilon}({z}_t^Y) = \epsilon_{\theta}({z}_t^Y) + (1 + s) \left( \epsilon_{\theta}({z}_t^Y, O_t^{Y \rightarrow X}) - \epsilon_{\theta}({z}_t^Y) \right). 
\end{equation}
Here, $s$ is the guidance scale.

\section{Experiments}
\label{sec:experiments}

\subsection{Experimental Settings}
\paragrapht{Dataset.}
Since there is no established benchmark for semantic image synthesis, we created a dedicated dataset featuring complex driving scenes with diverse structure-appearance pairs. Our dataset comprises 300 pairs collected from two widely used datasets: BDD100K~\cite{yu2020bdd100k} and NYUv2~\cite{Silberman:ECCV12}. Further details are provided in Appendix.

% For these datasets, we resize all images to $512\times 512$ during both training and inference. 
% Our approach enables T2I diffusion model with appearance transfer and structure control and we focus on the complex scenes (e.g., driving scenes). Therefore, we used BDD100K~\cite{yu2020bdd100k} dataset for two-stage training.
% BDD100K dataset consists of $100,000$ driving scene images under various conditions, of which $10,000$ images are paired with labels. 

% \paragrapht{Implementation Details.}
% \paragraphtt{Comparison Models.}
% We compare the performance of our model to the semantic image synthesis methods. We benchmarked our method against previous works including training-based model, ControlNet~\cite{zhang2023adding} + IP-Adapter~\cite{ye2023ip} and tuning-free model, FreeControl~\cite{mo2024freecontrol}, Cross-Image Attention~\cite{alaluf2024cross} and Ctrl-X~\cite{lin2024ctrl}. 

\paragrapht{Evaluation Metric.} Following previous studies~\cite{wang2022semantic, alaluf2024cross, lin2024ctrl, nam2024dreammatcher}, we evaluate our method using widely adopted metrics for (1) structural consistency, (2) appearance preservation, and (3) image quality. To assess structural consistency between the generated image and the given segmentation map, We use the self-similarity metric, denoted as Self-Sim ($\downarrow$)~\cite{tumanyan2022splicingvitfeaturessemantic}. To evaluate appearance preservation, we compute the CLIP image similarity denoted as $I_\mathrm{CLIP}$ ($\uparrow$)~\cite{radford2021learningtransferablevisualmodels} between the generated image and the exemplar image. To assess image quality, we calculate FID ($\downarrow$)~\cite{dhariwal2021diffusionmodelsbeatgans}. Additionally, to better capture local details, we calculated DINO-based metrics, including $I_{DINO} (\uparrow)$ and DINO[cls] loss $(\downarrow)$, which leverage self-supervised representation to assess fine-grained appearance fidelity. Further details on evaluation metrics are provided in Appendix.

\begin{figure}[t]
    \centering
    % \vspace{-10pt}
    \includegraphics[width=0.45\textwidth, height=3cm]{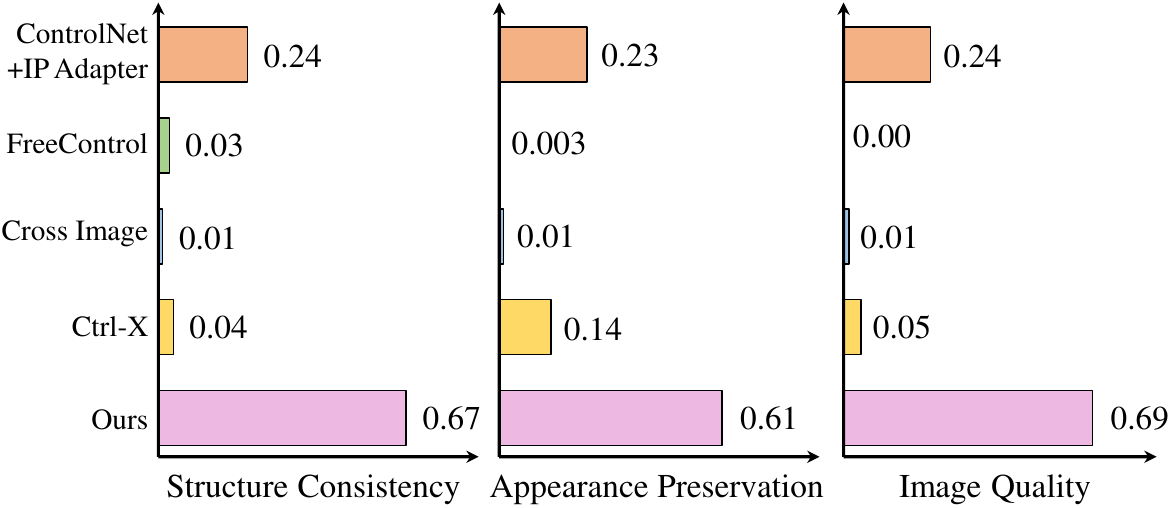} 
    % \vspace{-15pt}
    \caption{\textbf{User Study.}}
    \vspace{-15pt}
    \label{qual:human_eval}
\end{figure}

\paragrapht{User Study.}
We conducted a user study comparing our model to previous works~\cite{zhang2023adding, ye2023ip, mo2024freecontrol, alaluf2024cross, lin2024ctrl}. 45 participants evaluated the generated images based on structure preservation, appearance preservation and image quality. Additional details on the user study are in Appendix.

\subsection{Comparison}
Figure~\ref{qual:comp} and Table~\ref{tab:main} compare our method with prior models~\cite{zhang2023adding, ye2023ip, mo2024freecontrol, alaluf2024cross, lin2024ctrl}. ControlNet~\cite{zhang2023adding} + IP-Adapter~\cite{ye2023ip} achieves high structure preservation (Self-Sim), however, since IP-Adapter captures only global appearance, it often fails to preserve detailed appearance from the exemplar. Previous hand-crafted attention control methods~\cite{mo2024freecontrol, alaluf2024cross, lin2024ctrl} show suboptimal matching performance ($I_\mathrm{CLIP}$), leading to blurry and distorted outputs (FID) or disregarding structural guidance from target segmentation (Self-sim.). In contrast, AM-Adapter achieves superior structural consistency (Self-Sim), appearance preservation ($I_\mathrm{CLIP}$, DINO [cls] loss), and image quality (FID), demonstrating its effectiveness in preserving local details while closely reflecting the target structure. Figure~\ref{qual:human_eval} demonstrates that our method consistently surpasses prior works in structural consistency, appearance preservation and image quality. Additional results and applications are provided in Appendix.  
% cannot incorporate exemplars and relies solely on prompt-based descriptive guidance, which noticeably lowers the generation quality, especially in complex scenes such as driving scenarios. s
\subsection{Ablation Studies}
\label{sec:ablation_studies}
Figure \textcolor{NavyBlue}{12} in Appendix and Table~\ref{tab:abl_quan} demonstrate the effectiveness of each component. (I) ControlNeXt~\cite{peng2024controlnext} serves as the baseline, generating images without an exemplar. (a) and (II) show MasaCtrl~\cite{cao2023masactrl} , which directly copies the exemplar’s appearance by replacing the target keys and values, achieving high appearance scores.
% , which replaces the target keys and values with those of the exemplar, directly copying its appearance without modification. While it achieves high appearance scores($I_\mathrm{DINO}$, DINO [cls], $I_\mathrm{CLIP}$),
However, it entirely disregards semantic alignment, producing severely distorted outputs that completely fail to preserve target structure. (b) and (III) present DreamMatcher~\cite{nam2024dreammatcher}, which better maintains structure than (a) but remains ineffective for scene-level appearance transfer.
(c) and (IV) illustrate augmented self-attention without fine-tuning, performing best among hand-crafted attention methods(a, b) but still suffering from mismatches (e.g., truck matched to trees, lights matched to pedestrian).
% Unlike (I), (c) incorporates appearance, leading to a substantial improvement in appearance preservation. However, due to its reliance on implicit matching, it fails to correct mismatches (e.g., truck matched to trees, lights matched to pedestrian), resulting in distorted outputs. 
In comparison, (d) and (V) refine this with fine-tuning. While learning both matching and generation simultaneously, the unstable learning causes mismatched correspondences to remain unresolved and amplified during generation. To enhance matching, (e) and (VI) integrate the categorical matching cost to the augmented self-attention, but its matching performance remains limited, resulting in suboptimal outputs. In contrast, (f) and (VII) effectively disentangle matching from generation, maximizing both capabilities. This approach significantly improves overall image quality and instance-level details while demonstrating superior matching performance, as evident in transferred visual attributes such as vehicle color, and building structures. Additional results and detailed analysis are in the Appendix.

\section{Conclusion}
\label{sec:conclusion}
We introduce AM-Adapter for local appearance transfer by combining implicit matching costs from self-attention with categorical matching costs derived from segmentation maps. To disentangle structural guidance, generation, and matching processes, we employ a stage-wise training strategy. We also propose retrieval technique that optimizes matchable regions at inference. Extensive experiments demonstrate the effectiveness of our method, highlighting its robustness in challenging scenarios such as driving scenes, and several applications.
{
    \small
    \bibliographystyle{ieeenat_fullname}
    \bibliography{main}
}
\clearpage
\maketitlesupplementary
\appendix

In this material, Section~\ref{sec:appendix_impl} describes the implementation details of our experiments, while Section~\ref{sec:appendix_eval}
 elaborates the details of the evaluation process. Section~\ref{sec:appendix_results} presents additional results, including qualitative, quantitative analyses and user studies. Finally, in Section~\ref{sec:appendix_appl}, we demonstrate the versatility of our work through various applications. Section~\ref{sec:appendix_limitations} discusses the limitations of our work and provides a broader discussion. 
 
 % Finally, Section~\ref{sec:appendix_xl} presents experiments that utilize SDXL~\cite{podell2023sdxlimprovinglatentdiffusion} as the backbone.
 
% \section{Correction in Main paper}
% \label{sec:correction}
% In this section, we present corrections to errors in the main paper. Figure \textcolor{NavyBlue}{8} (b) depicts the results of augmented self-attention, not those of MasaCtrl~\cite{cao2023masactrl}. It is important to note that MasaCtrl employs key-value replacement, which replaces the key and value of the target with those of the exemplar, in contrast to augmented self-attention, which concatenates the key and value of the exemplar with those of the target. We provide further details and ablation studies in Section~\ref{sec:appendix_results}.
% 
\section{Implementation Details}
\label{sec:appendix_impl}
For all experiment, we used a single NVIDIA A6000 GPU. For training, we utilized 7K exemplar-target segmentation pairs from the BDD100K~\cite{yu2020bdd100k} dataset. During the first stage training, ControlNeXt~\cite{peng2024controlnext} was trained with a batch size of $2$ and a resolution of $512$, using a learning rate of $1e-5$, and the training process lasted approximately 9 hours. In the second stage, AM-Adapter was trained with a batch size of $1$ and a resolution of $512$ for $25,000$ steps, taking around $14$ hours. We used the same learning rate $1e-5$, with the AdamW~\cite{loshchilov2019decoupledweightdecayregularization} optimizer applied in both stages. 
We utilized the DDIM~\cite{song2020denoising} sampler for both inversion and sampling, with the number of timesteps $T$ set to $20$. The Augmented Self-Attention is applied across all self-attention layers of the UNet, denoted as $L\in[0, 9]$, while the AM-Adapter is applied to all self-attention layers except for the first block of the encoder, represented as $L \in [1, 9]$.  This approach avoids the structural conflict and prevents interference between the two effects, as the cross-normalized features from ControlNeXt~\cite{peng2024controlnext} are added after the first downsampling block. We set the guidance scale to $s=7.5$ for both text and matching guidance. 

% We concatenated all self-attention in every layer to the 

\section{Evaluation}
\label{sec:appendix_eval}
\subsection{Dataset}
\label{sec:appendix_eval_dataset}
Prior works in Semantic Image Synthesis~\cite{li2025controlnet, peng2024controlnext, zhao2024uni, zhang2023adding, park2019semantic, wang2022semantic, mou2024t2i, li2023gligen} and Exemplar-based Semantic Image Synthesis~\cite{ye2023ip, li2024photomaker, gal2023encoder, xiao2024fastcomposer, peng2024portraitbooth} have used different datasets for evaluation. Commonly, prior methods~\cite{lin2024ctrl} assemble image dataset from the web and hand annotated the condition such as segmentation, sketch or edge drawing. Since no established benchmark exists for Exemplar-based Semantic Image Synthesis, we constructed an evaluation dataset tailored to our task, featuring complex driving scenes with diverse structure-appearance pairs. Specifically, we evaluated our method on two commonly used driving scene datasets: BDD100K~\cite{yu2020bdd100k} and Cityscapes~\cite{cordts2016cityscapes}. To demonstrate generalization ability, we additionally evaluated our method on the NYUv2~\cite{Silberman:ECCV12} dataset, which is an indoor dataset. For evaluation, we randomly selected 300 segmentation maps each from BDD100K Cityscapes and NYUv2, resulting in a total of 900 segmentation maps.

Our goal is to achieve semantic-aware local appearance transfer in complex scenarios. As part of this effort, we propose a retrieval technique that automatically selects exemplars while maximizing matchable regions, as discussed in Section \textcolor{NavyBlue}{3.6} of the main paper. Therefore, the exemplar segmentation-image pairs in our evaluation dataset consist of two types: 300 pairs retrieved using 300 target segmentation maps and 300 pairs randomly selected.

% Our dataset comprises 300 pairs collected from two widely used datasets: BDD100K~\cite{yu2020bdd100k} and CityScapes~\cite{cordts2016cityscapes}. 

% For these datasets, we resize all images to $512\times 512$ during both training and inference. 
% Our approach enables T2I diffusion model with appearance transfer and structure control and we focus on the complex scenes (e.g., driving scenes). Therefore, we used BDD100K~\cite{yu2020bdd100k} dataset for two-stage training.
% BDD100K dataset consists of $100,000$ driving scene images under various conditions, of which $10,000$ images are paired with labels. 

\begin{figure*}
    \centering
    \includegraphics[width=\textwidth]{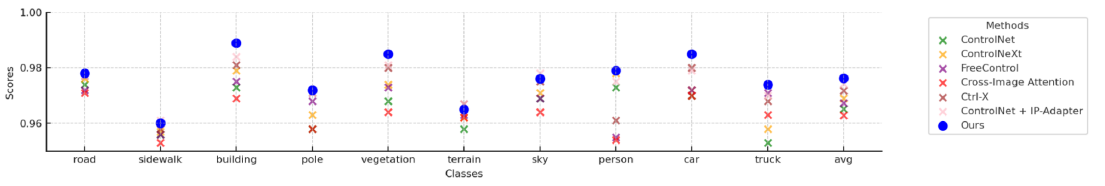} 
    \vspace{-20pt}
    \caption{\textbf{Object-wise CLIP Image Similarity per Class.}} %설명추가
    \vspace{-10pt}
    \label{supple:object_local_clip}
\end{figure*}

\subsection{Evaluation Metrics}
\label{sec:appendix_eval_metric}
For quantitative evaluation, we categorized our analysis into three perspectives: structure consistency, appearance preservation and image quality. Following previous studies~\cite{lin2024ctrl, parmar2024onestepimagetranslationtexttoimage}, we adopted Self-Sim.~\cite{tumanyan2022splicingvitfeaturessemantic} metrics to evaluate structure consistency. To evaluate appearance preservation, following prior works~\cite{nam2024dreammatcher}, we computed CLIP image similarity~\cite{radford2021learningtransferablevisualmodels}, denoted as $I_\mathrm{CLIP}$. For image quality assessment, we measured FID~\cite{dhariwal2021diffusionmodelsbeatgans}. Table \textcolor{NavyBlue}{1} of the main paper presents the results for these metrics. 

While our AM-Adapter outperforms other methods across all metrics, as noted in~\cite{christodoulou2024finding, tan2024evalalign, li2024genai, jiang2024genai, lin2025evaluating}, we emphasize that these metrics do not fully align with human preferences. This is because they extract either global features or small local features from the generated images and conditions (segmentation maps and exemplar RGB images) and calculate distances between these features. To address this limitation, we conducted the user study for a more reliable evaluation.

Furthermore, we evaluated three complementary metrics to address this limitation: object-wise local CLIP similarity, $I_{DINO}$ and DINO [cls] loss. The object-wise local CLIP similarity is  computed by categorizing objects and measuring the CLIP image similarity for each category individually. Since meaningful comparison requires the presence of corresponding objects in both the exemplar and result images, we restrict our analysis to the top 10 most frequently occurring object classes in the BDD100K~\cite{yu2020bdd100k} dataset. Despite this constraint, our method consistently demonstrates superior robustness across all classes compared to other baseline models. 
Additionally, we employ DINO, which is known for its supervised learning capability and its ability to capture fine-grained local details, to measure image similarity. Our approach achieves the best performance in this evaluation. Table \textcolor{NavyBlue}{1} of the main paper demonstrates the results for $I_\mathrm{DINO}$ and DINO [cls] loss, and Figure~\ref{supple:object_local_clip} shows the results of object-wise local CLIP similarity across categories in BDD100K dataset.

% \input{supple/tables/dino_table}
% Therefore, we propose two complementary metrics to address this limitation: $Local-CLIP$ and $Local-DINO$. 
% ... 확정되면 마저. 

\subsection{User Study Details}
\label{sec:appendix_eval_userstudy}
\begin{figure*}
    \centering
    \includegraphics[width=\linewidth]{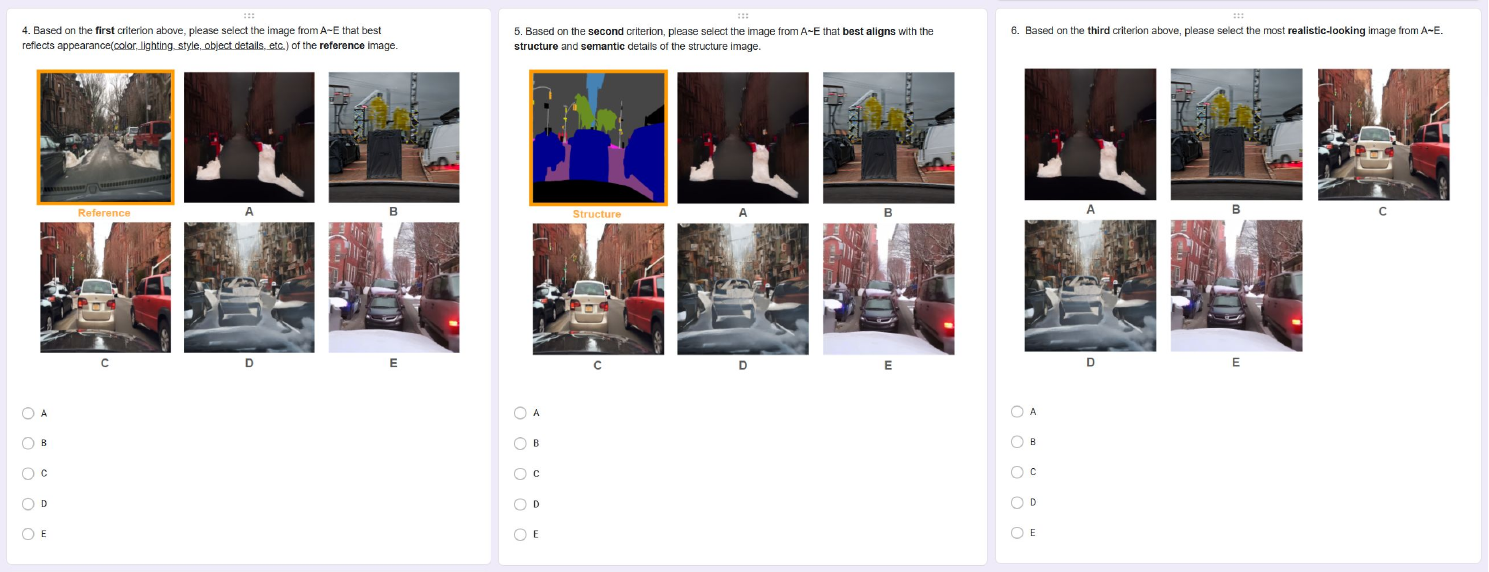}  
    \vspace{-10pt}
    \caption{\textbf{An Example of a User Study Comparing AM-Adapter with Previous Methods.} For structure preservation, we provide the target segmentation map and generated images from different methods, ControlNet~\cite{zhang2023adding} + IP-Adapter~\cite{ye2023ip}, FreeControl~\cite{mo2024freecontrol}, Cross-Image Attention~\cite{alaluf2024cross}, Ctrl-X~\cite{lin2024ctrl} and AM-Adapter. For appearance preservation, we provide the exemplar image and the generated images from those methods. For image quality, we compare solely the generated images. For a fair comparison, we randomly select the sample generated from exemplar-segmentation pairs from a large pool. }
    \vspace{-10pt}
    
    \label{supp:user_study_ex}
\end{figure*}

Figure~\ref{supp:user_study_ex} presents an example question from the user study. We conducted a human evaluation study comparing AM-Adapter and previous works~\cite{zhang2023adding, ye2023ip, mo2024freecontrol, alaluf2024cross, lin2024ctrl} in terms of structure consistency, appearance preservation, and image quality. For structure consistency, we provided the target segmentation map and the generated images from different methods, and users were asked to select which method better represents the semantic structure in the target segmentation map. For appearance preservation, we provided the exemplar image and the generated images from different methods, and participants were asked to choose which generated image better captures the appearance in the exemplar. Lastly, participants were shown only the generated images and asked to select the one that achieved the highest image quality. A total of 45 participants responded to 18 questions. For a fair comparison, we sampled generated images from a large pool sharing the same exemplar image for three different methods to ensure intra-rater reliability. Figure \textcolor{NavyBlue}{9} in the main paper summarizes the results, showing that our model outperforms others across all three criteria.

\begin{figure}[t]
    \centering
    % \vspace{-10pt}
    \includegraphics[width=0.48\textwidth]{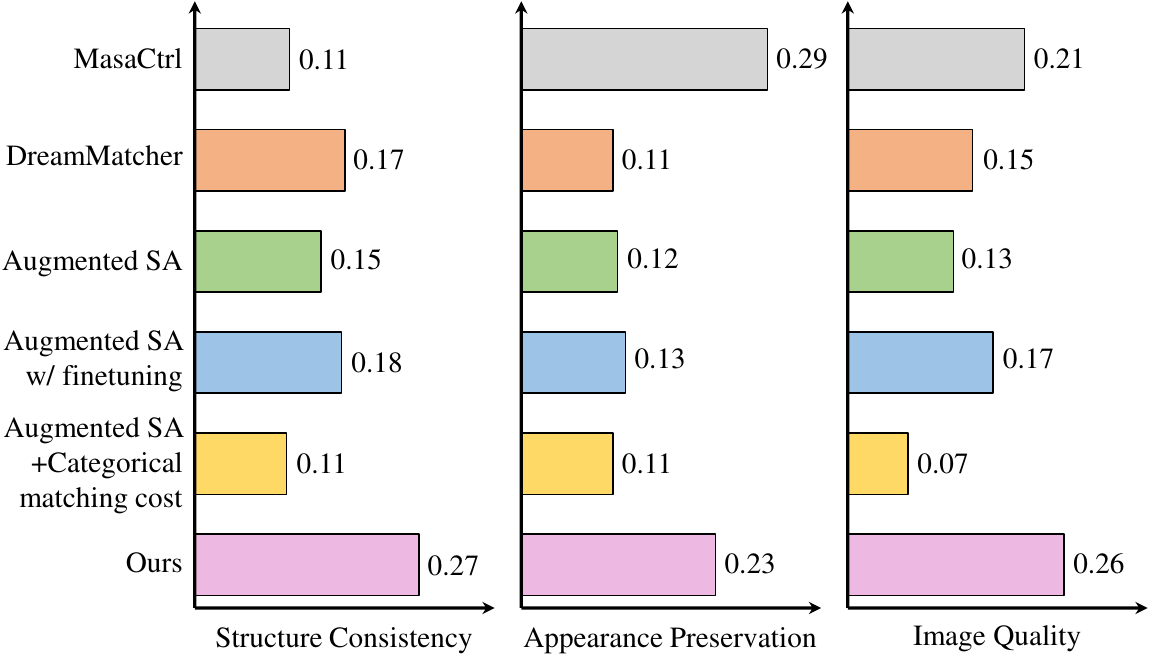} 
    \vspace{-15pt}
    \caption{\textbf{User Study on Ablations.}}
    \vspace{-15pt}
    \label{supple:user_study_abl}
\end{figure}

\begin{figure*}
    \centering
    \includegraphics[width=\linewidth]{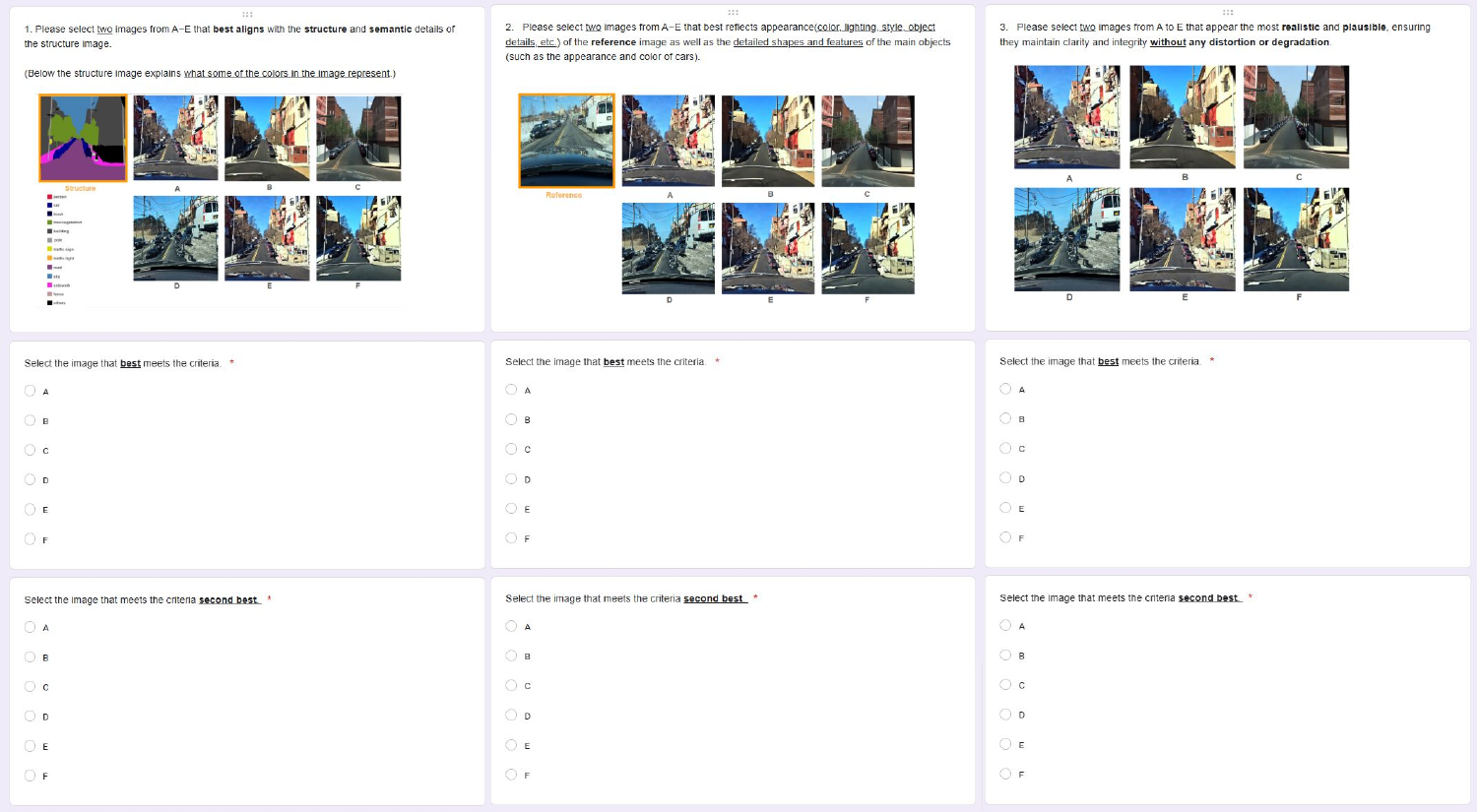} 
    \vspace{-10pt}
    \caption{\textbf{An Example of a User Study Comparing AM-Adapter with Ablation Studies.} For structure preservation, we provide the target segmentation map and generated images from different methods, key-value replacement (MasaCtrl~\cite{cao2023masactrl}), DreamMatcher~\cite{nam2024dreammatcher}, augmented self-attention, augmented self-attention w/ finetuning, augmented self-attention + categorical matching cost, and \textbf{AM-Adapter}. For appearance preservation, we provide the exemplar image and the generated images from those methods. For image quality, we compare solely the generated images. For a fair comparison, we randomly select the sample generated from exemplar-segmentation pairs from a large pool. }
    \vspace{-10pt}
    
    \label{supp:user_study_abl_ex}
\end{figure*}

Figure~\ref{supp:user_study_abl_ex} shows an example question from the user study comparing AM-Adapter with its individual components. The evaluation was conducted using the same criteria of structure preservation, appearance preservation and image quality to assess the role of each component in the overall performance. A total of 33 participants responded to 24 questions. For a fair comparison, generated images were samples from a shared pool associated with the same exemplar image and same target segmentation map across all methods. The results, summarized in Figure~\ref{supple:user_study_abl}, show that AM-Adapter achieves a balanced and consistently high performance across all three criteria.  

\section{Additional Results}
\label{sec:appendix_results}

\subsection{Qualitative Results}
\label{sec:appendix_results_qual}
\begin{figure*}
    \centering
    \includegraphics[width=\linewidth]{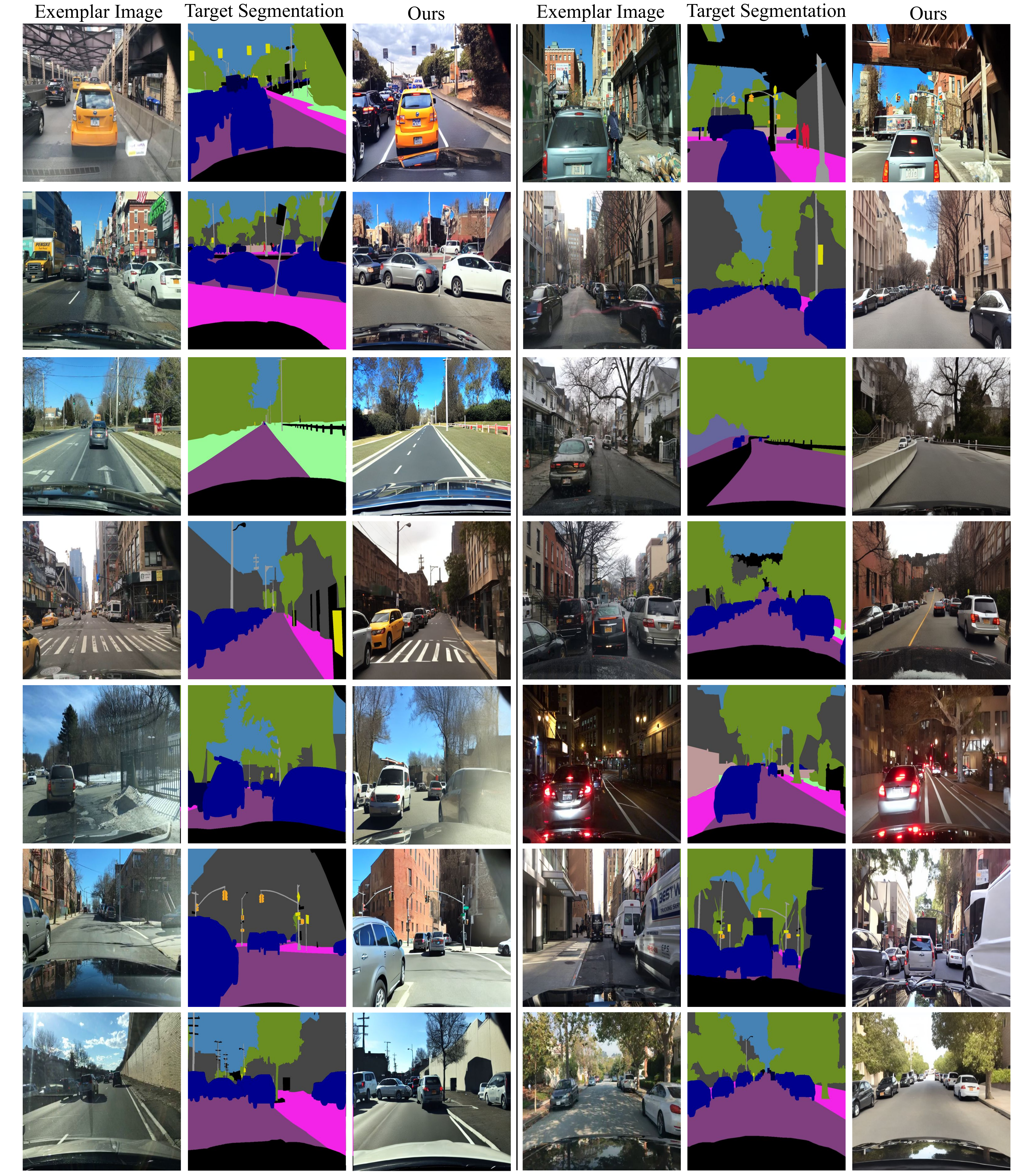} 
    \vspace{-10pt}
    \caption{\textbf{Additional Qualitative Results of AM-Adapter. } Visualization of results generated by \textbf{AM-Adapter (Ours)} across various scenarios.}
    \vspace{-10pt}
    
    \label{supp:additional_qual_ours}
\end{figure*}

\begin{figure*}
    \centering
    \includegraphics[width=\linewidth]{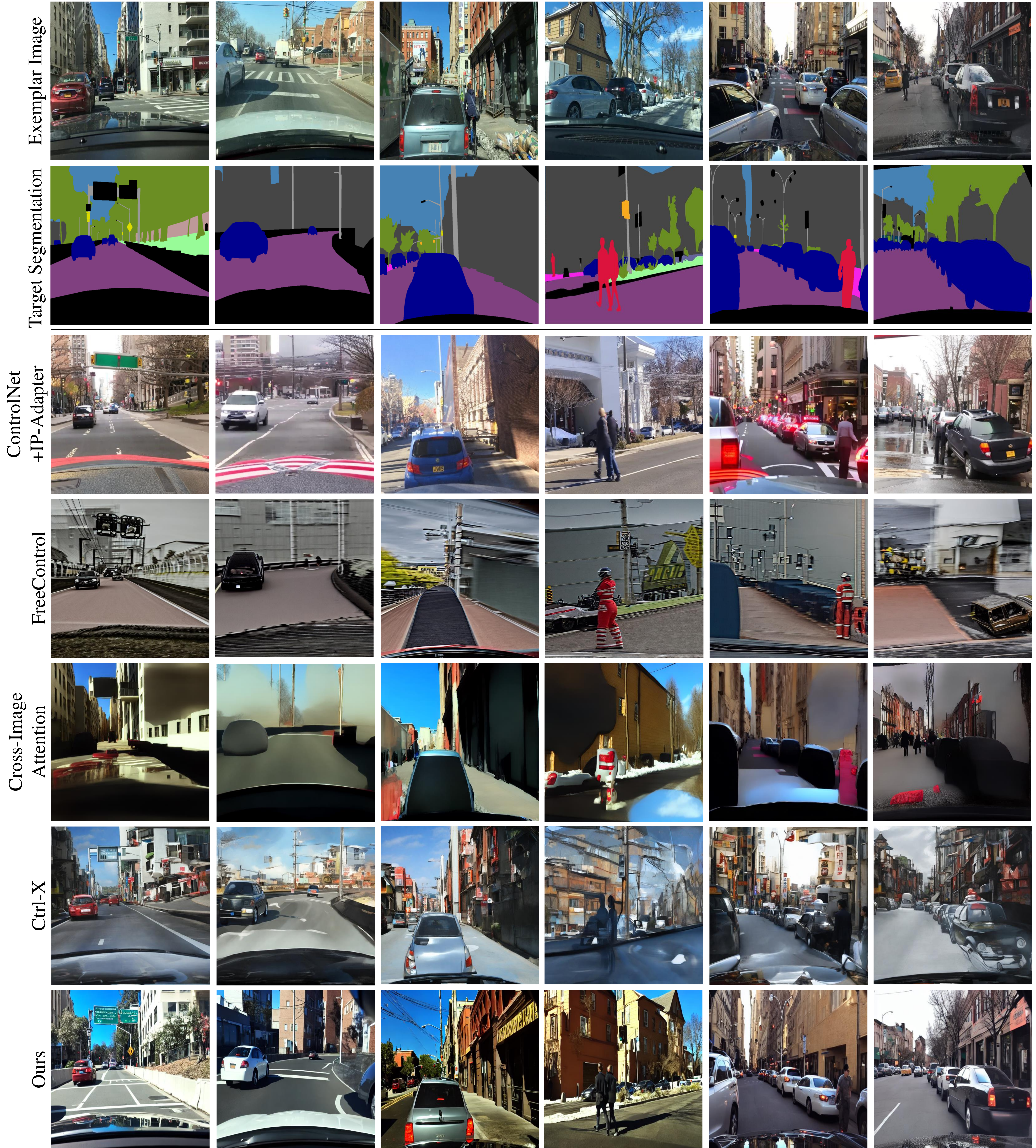} 
    \vspace{-10pt}
    \caption{\textbf{Additional Qualitative Comparison on BDD100K~\cite{yu2020bdd100k} Dataset.}}
    \vspace{-10pt}
    \label{supp:additional_qual_comp}
\end{figure*}

Figure~\ref{supp:additional_qual_ours} shows additional qualitative results of AM-Adapter. We present more qualitative results comparing our method with others, including ControlNet~\cite{zhang2023adding} + IP-Adapter~\cite{ye2023ip}, FreeControl~\cite{mo2024freecontrol}, Cross-Image Attention~\cite{alaluf2024cross}, and Ctrl-X~\cite{lin2024ctrl}, in Figure~\ref{supp:additional_qual_comp}, which further demonstrates the effectiveness of our method.

\subsection{Ablation Study Analysis}
\label{sec:appendix_results_abl}

\begin{figure*}%[t]
    \centering
    % \vspace{-20pt}
    \includegraphics[width=\textwidth]{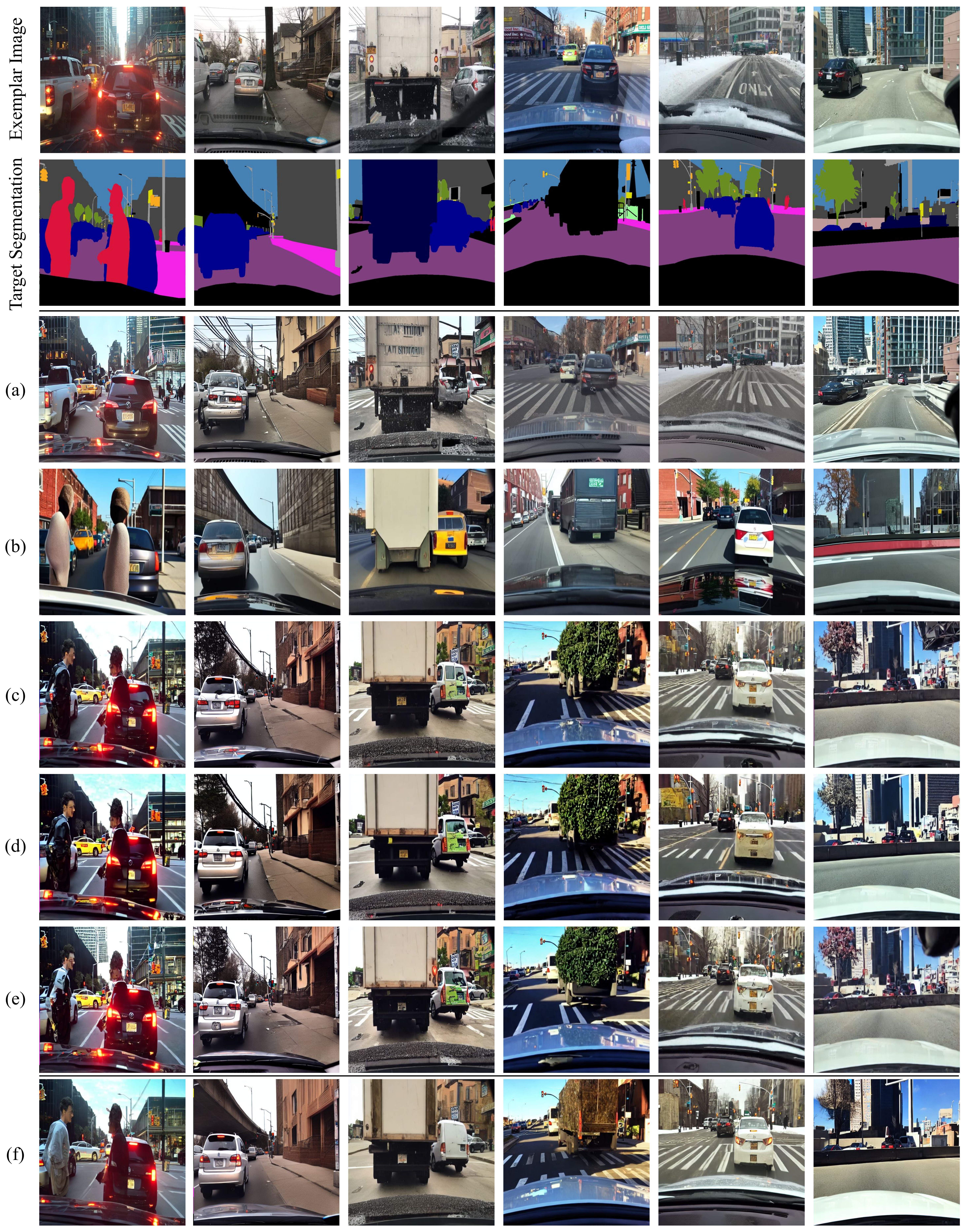} 
    \vspace{-15pt}
    \caption{\textbf{Ablation Study on Individual Components:} (a) ControlNeXt~\cite{peng2024controlnext} + MasaCtrl~\cite{cao2023masactrl}, (b) ControlNeXt~\cite{peng2024controlnext} + DreamMatcher~\cite{nam2024dreammatcher}, (c) ControlNeXt~\cite{peng2024controlnext} + Augmented Self-Attention, (d) (c) + w/ Fine-tuning, (e) (c) + Categorical Matching Cost, (f) \textbf{AM-Adapter (Ours)}.}
    \vspace{-15pt}
    \label{qual:abl}
\end{figure*}

\begin{figure*}[t]
    \centering
    \includegraphics[width=\textwidth]{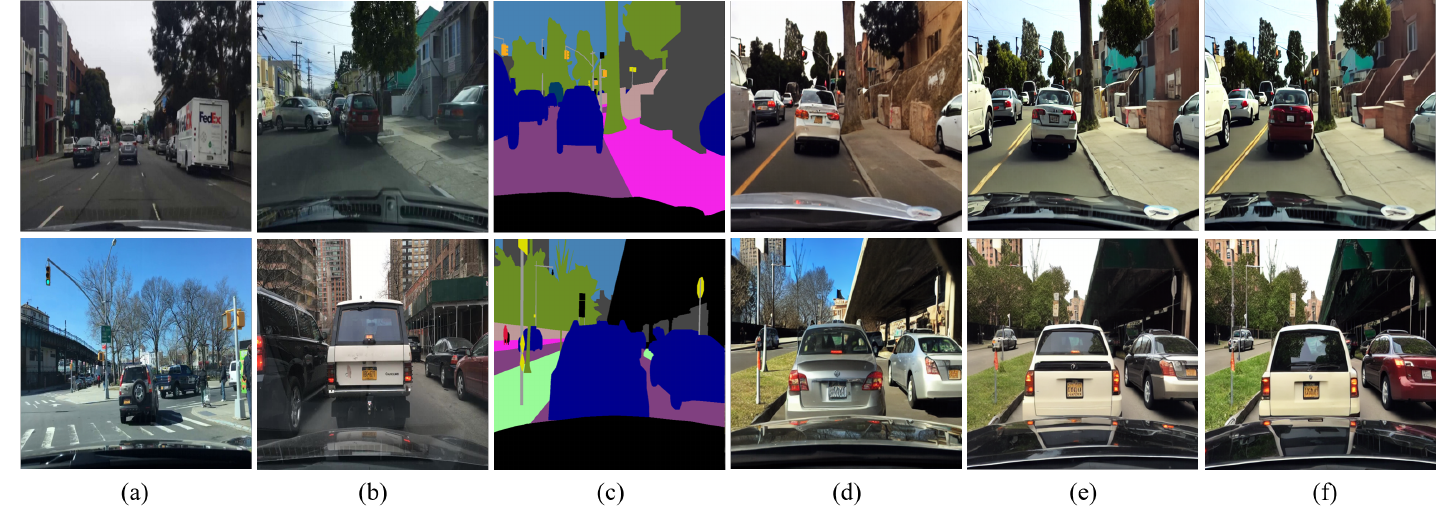} 
    \vspace{-20pt}
    \caption{\textbf{Ablation on Inference:} (a) random exemplar image, (b) retrieved exemplar image, (c) target segmentation map with desired structure, results generated (d) without retrieval or matching guidance, (e) with retrieval but without matching guidance, and (f) with both retrieval and matching guidance.}
    \vspace{-10pt}
    
    \label{supple:retrieval_abl}
\end{figure*}

Figure~\ref{qual:abl},
Figure~\ref{supple:retrieval_abl}, 
Table \textcolor{NavyBlue}{2} in the main paper and Table~\ref{supple:quan_inference_abl} summarize the following ablation study.

\paragrapht{MasaCtrl vs. DreamMatcher vs. Augmented Self-Attention.} As shown in  Figure~\ref{qual:abl} and Table \textcolor{NavyBlue}{2}, we conducted a comparative analysis of the Augmented Self-Attention against two representative hand-crafted attention control methods, MasaCtrl~\cite{cao2023masactrl} and DreamMatcher~\cite{nam2024dreammatcher}, which focus on implicit matching within self-attention mechanisms. MasaCtrl replaces the key and value of the synthesized target image with those of the exemplar image, while DreamMatcher enhances implicit matching by leveraging diffusion features for improved correspondence.

MasaCtrl heavily relies on implicit matching in the self-attention module, often resulting in inaccurate appearance transfers to semantically misaligned regions, thereby disrupting the target structure that aligns with the structural conditions. Furthermore, key-value replacement cannot generate new elements that are absent in the exemplar, as it discards the original keys and values from the target and replaces them with those from the exemplar. 
This limitation causes key-value replacement (MasaCtrl) to exhibit high $I_\mathrm{CLIP}$ and $I_\mathrm{DINO}$ and low DINO [cls] loss, as it discards the original key-value pairs of the target and copies and pastes the exemplar's appearance into the target image based on incorrect matching. This is further demonstrated by the Self-Sim.~\cite{tumanyan2022splicingvitfeaturessemantic} of MasaCtrl in Table \textcolor{NavyBlue}{2} and the structural consistency results from the user study Figure \textcolor{NavyBlue}{9}, which show that key-value replacement disrupts the structural consistency between generated images and the given semantic conditions. Notably, the metric of exemplar image in Table \textcolor{NavyBlue}{2} reflects cases where the same appearance exemplar is used, which naturally leads to very high $I_\mathrm{CLIP}$ scores due to the identical appearance. Similarly, MasaCtrl achieves high scores even when matching fails, as it transfers all appearance information from the exemplar to the target.

In contrast, as demonstrated in Figure~\ref{qual:abl} and Table \textcolor{NavyBlue}{2}, DreamMatcher better preserves the target structure by leveraging improved semantic matching and using intermediate diffusion features to align the exemplar’s appearance with the target image. However, its appearance preservation is highly dependent on the exemplar image, as relying solely on hand-crafted matching with diffusion features is insufficient for establishing accurate correspondence in content-rich images, such as driving scenes, which require precise matching.

As a result, to achieve our goal of accurate transfer in complex scenes, we adopt Augmented Self-Attention as our baseline, which is outlined in Section \textcolor{NavyBlue}{3.3} of the main paper. As shown in Figure~\ref{qual:abl} and Table \textcolor{NavyBlue}{2}, Augmented Self-Attention selectively incorporates keys and values from both the exemplar and the target, effectively preserving the desired target structure while transferring the exemplar's local appearance. Due to this selectivity, $I_\mathrm{CLIP}$ is lower compared to key-value replacement, which indiscriminately transfers the entire appearance information from the exemplar. However, Augmented Self-Attention demonstrates a superior human evaluation score compared to MasaCtrl~\cite{cao2023masactrl} and DreamMatcher~\cite{nam2024dreammatcher}, indicating that it concurrently achieves appearance transfer and structural consistency.

\paragrapht{Impact of Fine-tuning.}
Nevertheless, as discussed in Section \textcolor{NavyBlue}{3.3} in the main paper, relying solely on the implicit matching of the Augmented Self-Attention still leads to the mismatches. The most straightforward approach to address this issue is fine-tuning the model. However, as shown in Figure~\ref{qual:abl} (c) and (d) and discussed in Section \textcolor{NavyBlue}{4.3}, it is shown that fine-tuning degrades detailed appearance preservation, as it requires learning both generation and matching simultaneously, which leads to unstable training and overfitting. This is further demonstrated in the (V) in  Table \textcolor{NavyBlue}{2}, (d) in Figure~\ref{qual:abl}. 

\paragrapht{Categorical Matching Cost vs. Learnable Matching Cost (AM-Adapter).}
To concurrently achieve matching and generation, we refine the implicit matching with semantic awareness using segmentation maps in a data-driven manner, rather than fine-tuning the entire model. In Section \textcolor{NavyBlue}{3.4} of the main paper, we compare categorical matching cost with AM-Adapter. (e) and (f) in Figure~\ref{qual:abl} further demonstrate the effectiveness of AM-Adapter compared to categorical matching cost.

\begin{figure*}[t]
    \centering
    \includegraphics[width=\textwidth]{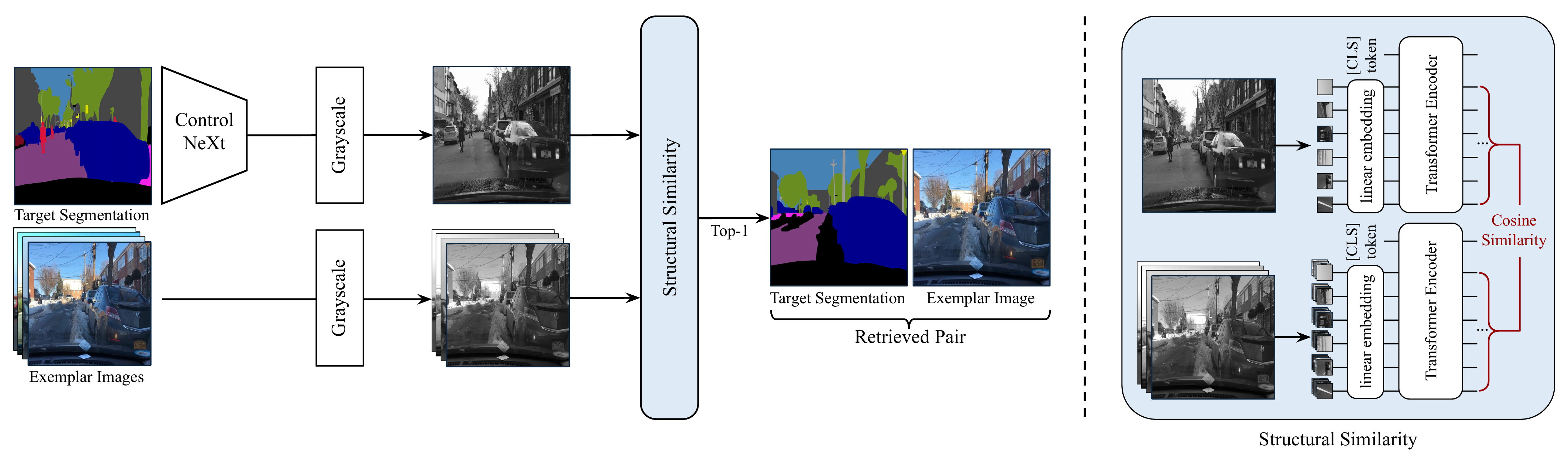} 
    \vspace{-20pt}
    \caption{\textbf{Details of the Retrieval-based Inference.}}
    \vspace{-10pt}
    
    \label{supple:retrieval_detail}
\end{figure*}

\begin{figure*}[t]
    \centering
    \includegraphics[width=\textwidth]{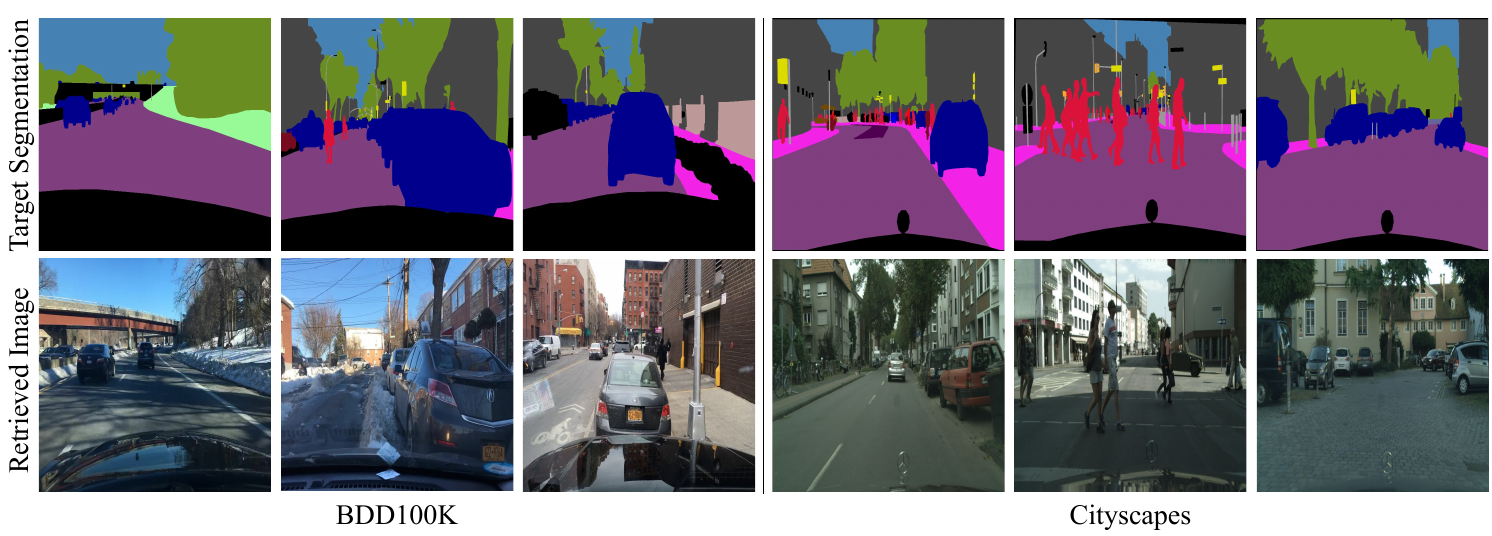} 
    \vspace{-20pt}
    \caption{\textbf{Retrieval Examples.} During inference, our retrieval technique selects the exemplar image that exhibits the highest structural similarity to the target segmentation map.}
    \vspace{-10pt}
    
    \label{supple:retrieval_ex}
\end{figure*}

\paragrapht{Inference.}
\label{sec:appendix_abl_inference}
Figure~\ref{supple:retrieval_abl} and Table~\ref{supple:quan_inference_abl} illustrate the effects of our proposed retrieval technique and matching guidance during inference. (a) represents a randomly selected exemplar image, and (b) represents a retrieved exemplar image. (c) and (I) display the results when a random exemplar image is used as input without retrieval. In contrast, (d) and (II) utilize an exemplar image that is structurally most similar to the target condition, resulting in higher appearance preservation compared to (a). Notably, this retrieval-based approach results in a significant increase in $I_\mathrm{CLIP}$ scores, increasing the matchable regions and transferring more appearance information from the exemplar. Finally, (e) and (III) show the results of applying matching guidance using classifier-free guidance~\cite{ho2022classifierfreediffusionguidance}, highlighting the effectiveness of matching guidance in achieving accurate results. Figure~\ref{supple:retrieval_detail} presents the detailed mechanism of how the retrieval process operates and Figure~\ref{supple:retrieval_ex} illustrates retrieved exemplar images from target segmentation maps obtained by our retrieval technique.

\subsection{Additional Analysis}
\begin{figure*}
    \centering
    \includegraphics[width=\textwidth, height=5.5cm]{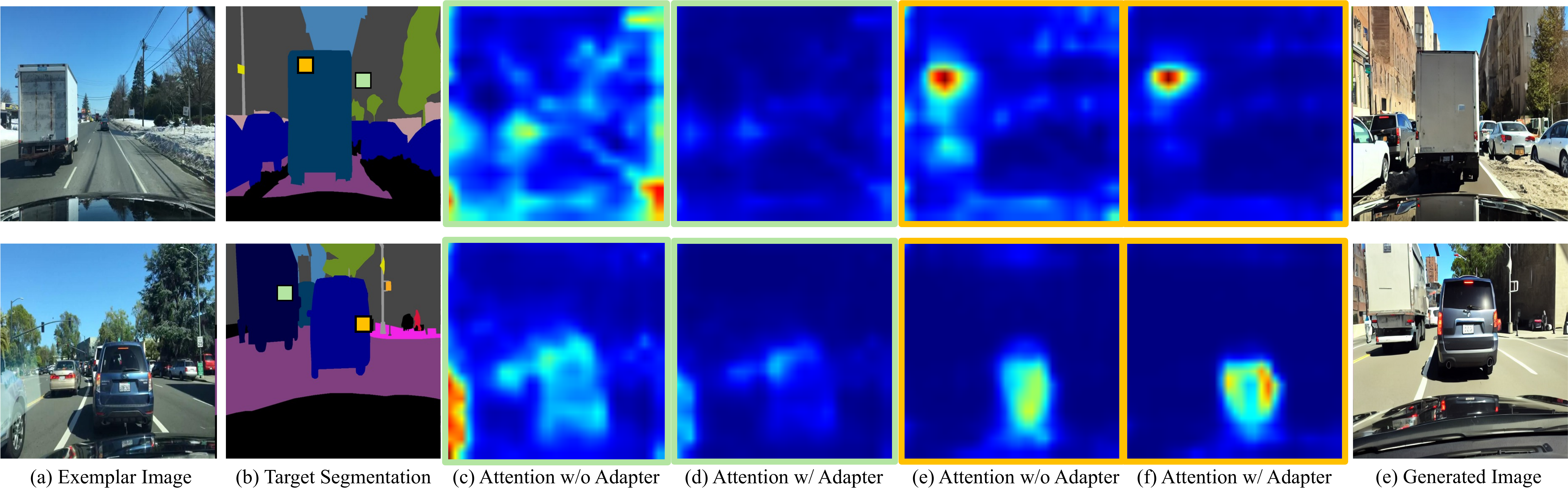} 
    \vspace{-20pt}
    \caption{\textbf{Additional Attention Visualization:} (a) Exemplar image with desired appearance, (b) target segmentation with desired structure, and (g) generated image. \textcolor{green}{Green} and \textcolor{orange}{orange} markers in (b) indicate query points. (c) and (d) show the augmented self-attention map $Q_t^Y (K_t^X)^T$ from the {green} marker, before and after applying AM-Adapter, respectively. (e) and (f) show the augmented self-attention map $Q_t^Y (K_t^X)^T$ from the {orange} marker, before and after applying AM-Adapter, respectively.}
    \vspace{-10pt}
    \label{supple:attn_vis}
\end{figure*}

Figure~\ref{supple:attn_vis} presents the additional examples of attention visualizations before and after applying the AM-Adapter, further highlighting the effectiveness of our model. The green markers indicate objects that are absent in the exemplar, while the orange markers denote objects that are present in the exemplar. 

For instance, in the first row, the green marker highlights a query point corresponding to a `building' that is absent in the exemplar. After applying the AM-Adapter, the attention associated with unrelated regions is significantly suppressed. The orange marker, on the other hand, indicates a query point on the top of a `trailer'. After applying AM-Adapter, the attention becomes more localized, concentrating more effectively on the relevant regions. In the second row, the green marker represents a query point corresponding to a `truck' that does not appear in the exemplar. After the AM-Adapter is applied, mismatches in the attention are reduced, further aligning the attention with the target structure. Similarly, the orange marker identifies a query point near the right rear light of a `vehicle'. With the adapter applied, the attention becomes more localized, exhibiting enhanced focus on the intended area. 

% \input{author-kit-CVPR2025-v3.1-latex-/supple/tables/quan_kv}
% \subsection{User study}
% Figure~\ref{supple:user_study_abl} summarizes a human evaluation study to compare AM-Adapter with ablation studies of its individual components focusing on the criteria of structure consistency, appearance preservation and image quality, as used in the user study detailed in the main paper. A total of 33 participants each responded to 24 questions. For a fair comparison, we sampled generated images from a large pool, using the same exemplar-target segmentation pairs for different components. Figure~\ref{supp:user_study_abl_ex} presents an example question of the user study.

% \begin{table}[t]
% % \small
% % \centering
% \resizebox{\columnwidth}{!}{
%     \begin{tabular}{l|l|ccc|}
%     \toprule
%     & Component & $I_{DINO}$ $\uparrow$  & $I_{CLIP} \uparrow$ &FID$\downarrow$   \\
%     \midrule
%     (I) & AM-Adapter  & 0.723 & 0.741 & 79.05 \\
%     (II) & (I) + Retrieval & \textbf{0.787} & \underline{0.814} & \underline{77.15} \\ 
%     \bluerow (III) & (II) + Matching Guidance \textbf{(Ours)} & 
%     \underline{0.786} & \textbf{0.819} &  \textbf{75.89} \\
%     \bottomrule
%     \end{tabular}
% }
% \caption{\textbf{Ablation on inference.}}
% \label{supple:quan_inference_abl}
% \end{table}

\begin{table}[t]
% \small
% \centering
\resizebox{\columnwidth}{!}{
    \begin{tabular}{l|l|ccc}
    \toprule
    & Component & Self-Sim. $\downarrow$  & $I_\mathrm{CLIP} \uparrow$ &FID$\downarrow$   \\
    \midrule
    (I) & AM-Adapter  & 0.043 & 0.741 & 79.05 \\
    (II) & (I) + Retrieval & \underline{0.041} & \underline{0.814} & \underline{77.15} \\ 
    (III) & (II) + Matching Guidance \textbf{(Ours)} & 
    \textbf{0.041 }& \textbf{0.819} &  \textbf{75.89} \\
    \bottomrule
    \end{tabular}
}
\caption{\textbf{Ablation Study on Inference.}}
\label{supple:quan_inference_abl}
\end{table}

% \section{Generalization}
% \label{sec:appendix_general}
\subsection{Generalization}
\begin{figure*}[t]
    \centering
    \includegraphics[width=\textwidth]{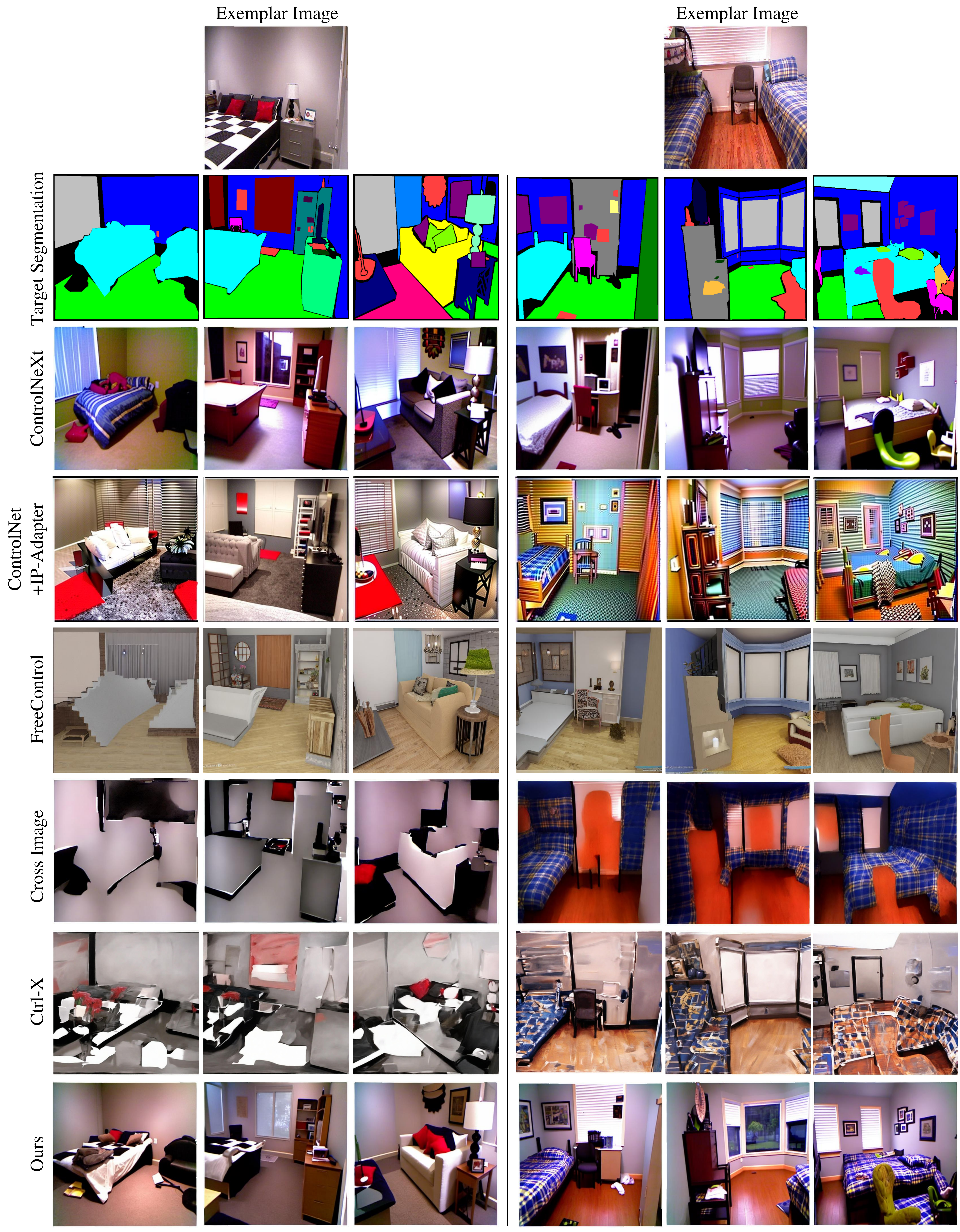} 
    \vspace{-20pt}
    \caption{\textbf{Additional qualitative results on NYUv2~\cite{Silberman:ECCV12} dataset.}}
    \vspace{-10pt}
    
    \label{supple:nyu_eval}
\end{figure*}

Our method is not limited to driving scenes. To demonstrate its generalizability across domains, we additionally evaluated it on NYUv2~\cite{Silberman:ECCV12}, an indoor dataset with 40 classes, as shown in Figure~\ref{supple:nyu_eval}.

% \input{author-kit-CVPR2025-v3.1-latex-/supple/tables/quan_local_metric}
% \input{author-kit-CVPR2025-v3.1-latex-/supple/tables/quan_local_abl}

% \subsection{Robustness to Exemplar}

% \input{supple/figures/cross_domain_setting}
% \input{supple/figures/exemplar_robustness}
%설명 + 피겨 수정?
\section{Applications}
\label{sec:appendix_appl}
\subsection{Controllable One-to-One Appearance Transfer with User Guidance}
\begin{figure*}
    \centering
    \includegraphics[width=\linewidth]{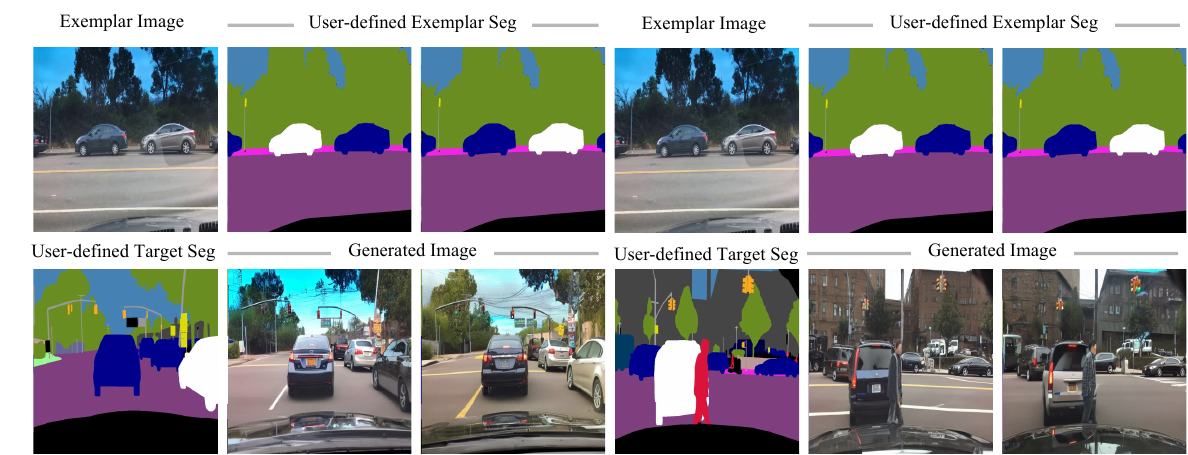} 
    \vspace{-10pt}
    \caption{\textbf{Application: Controllable One-to-One Matching with User Guidance.} The AM-Adapter enforces one-to-one mapping in a many-to-many setting, enabling controlled appearance transfer. The white regions in the exemplar and target segmentations indicate the source and destination objects, respectively, ensuring accurate appearance mapping.
    }
    % (a) Target segmentation map with the desired structure, (b) an exemplar image with rainy conditions, (c) the results generated under the exemplar condition in (b), (d) an exemplar image with sunny conditions, (e) the results generated under the exemplar condition in (d), (f) an exemplar image with night conditions, (g) the results generated under the exemplar condition in (f).}
    %\vspace{-10pt}
    
    \label{supp:application_control}
\end{figure*}

As illustrated in Figure~\ref{supp:application_control}, the AM-Adapter can enforce one-to-one mapping within a many-to-many setting, allowing precise control over appearance transfer. In the user-defined exemplar segmentation, the white regions indicate source vehicles whose appearance is explicitly designated for transfer. In the user-defined target segmentation, the white regions represent destination objects that will receive the transferred appearance. As a result, the appearance of the selected source vehicle in the exemplar is accurately mapped to the corresponding destination vehicle in the target segmentation.

\subsection{Segmentation-Guided Image Editing}
\begin{figure*}
    \centering
    \includegraphics[width=\linewidth]{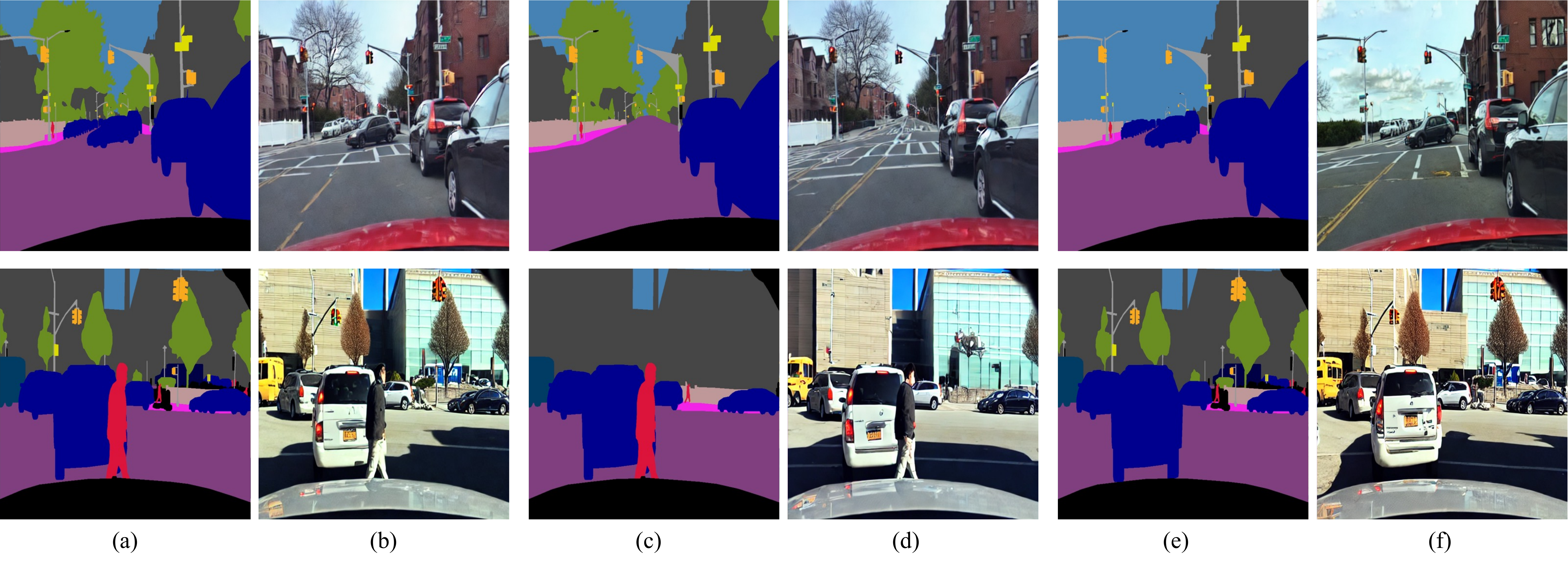} 
    \vspace{-20pt}
    \caption{\textbf{Application: Object Removal by Segmentation-Based Image Editing.} (a) Original target segmentation with desired structure, (b) generated image given the target segmentation map (a), (c) edited target segmentation map, (d) generated image given the target segmentation map (c), (e) edited target segmentation map, (f) generated image given the target segmentation map (e).}
    \vspace{-10pt}
    
    \label{supp:application_seg_edit}
\end{figure*}

\begin{figure*}
    \centering
    \includegraphics[width=\linewidth]{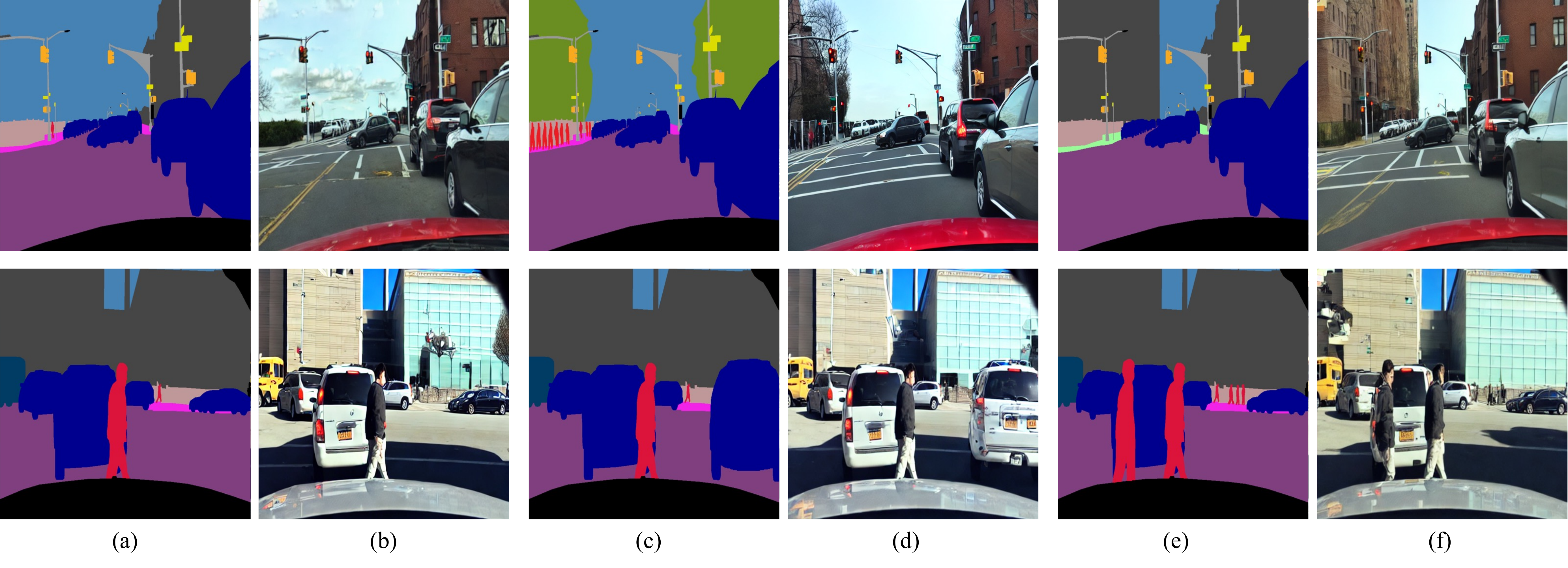} 
    \vspace{-20pt}
    \caption{\textbf{Application: Object Additon by Segmentation-Based Image Editing.} (a) Edited target segmentation with desired structure, (b) generated image given the target segmentation map (a), (c) edited target segmentation map, (d) generated image given the target segmentation map (c), (e) edited target segmentation map, (f) generated image given the target segmentation map (e).}
    \vspace{-10pt}
    
    \label{supp:application_seg_edit_add}
\end{figure*}

Figure~\ref{supp:application_seg_edit} and Figure~\ref{supp:application_seg_edit_add} illustrate the results of AM-Adapter to segmentation-based editing applications, specificallly object removal and addition, respectively. These results showcase the versatility of AM-Adapter in generating high-quality images while adhering to modified semantic structures.

In Figure~\ref{supp:application_seg_edit}, we demonstrate segmentation-based editing by removing objects from the segmentation maps. (a) represents the original segmentation map, while (b) shows the generated image by AM-Adapter following the target structure of (a). (c) and (e) are edited segmentation maps derived from (a) with specific modifications, such as modifying or removing the colors of specific instances within the map to adjust the semantic information. For example, in the first row, (c) removes the vehicles in the center, while (e) removes the buildings on the left. In the second row, (c) eliminates the traffic lights and trees, while (e) removes the person. (d) is the generated image by AM-Adapter following the target structure of (c), while (f) is the generated image by AM-Adapter following the target structure of (e). 

Figure~\ref{supp:application_seg_edit_add}, in contrast, showcases segmentation-based editing through object addition in the segmentation maps. (a) depicts the edited segmentation maps from Figure~\ref{supp:application_seg_edit} where specific objects were removed, forming the basis for subsequent augmentation, and (b) presents the image generated based on (a). (c) and (e) illustrate segmentation maps augmented by introducing additional semantic instances to enrich the scene. For example, in the first row, (c) introduces pedestrians into the left side of the scene, while (e) adds buildings to the background. In the second row, (c) incorporates a vehicle into the scene, while (e) includes pedestrians. The resulting images, generated based on these augmented segmentation maps, are depicted in (d) and (f), highlighting the model's capability to seamlessly integrate new objects into the scene while preserving global consistency. 

% These results demonstrate the versatility of our method, highlighting its applicability to downstream tasks such as semantic editing~\cite{luo2023siedobsemanticimageediting}. 
The results shown in Figure~\ref{supp:application_seg_edit} and Figure~\ref{supp:application_seg_edit_add} highlight the versatility of AM-Adapter in addressing diverse semantic editing tasks, including object removal and addition. The method effectively synthesizes images that reflect both the structural modifications and the appearance of exemplar, showcasing its potential for downstream applications such as semantic image editing~\cite{luo2023siedobsemanticimageediting}. These findings further validate AM-Adapter's contribution to advancing exemplar-based semantic image synthesis.

\subsection{Image-to-Image Translation}
\begin{figure*}
    \centering
    \includegraphics[width=\linewidth]{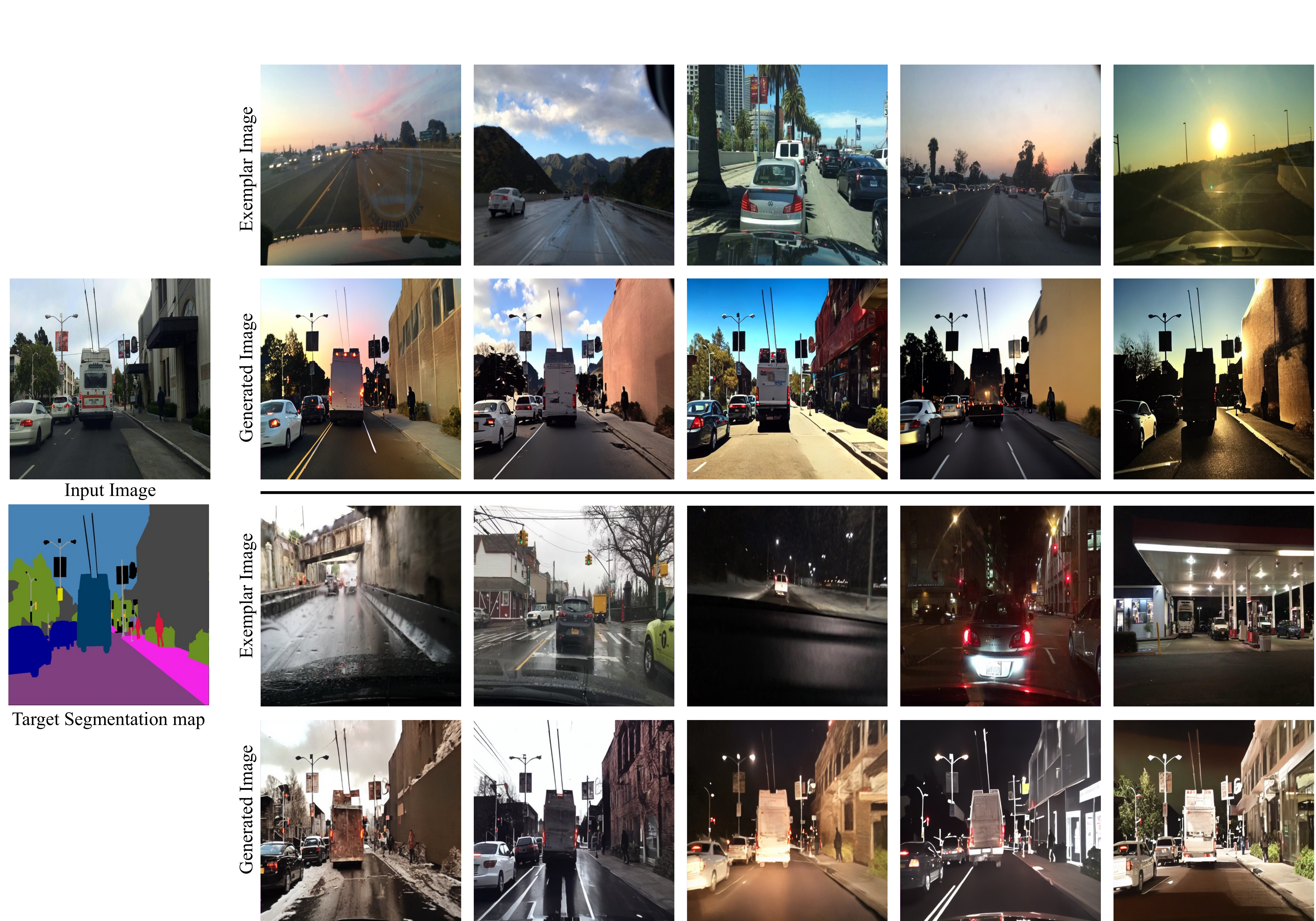} 
    \vspace{-10pt}
    \caption{\textbf{Application: Image-to-Image Translation.} The first and third row represent exemplar images reflecting various weather conditions and times of day, including various categories such as cloudy days, sunset hours, sunny, night, and rainy conditions. The second and fourth rows depict the resulting images that incorporate the structure of the target segmentation along with each exemplar image.
    }
    % (a) Target segmentation map with the desired structure, (b) an exemplar image with rainy conditions, (c) the results generated under the exemplar condition in (b), (d) an exemplar image with sunny conditions, (e) the results generated under the exemplar condition in (d), (f) an exemplar image with night conditions, (g) the results generated under the exemplar condition in (f).}
    %\vspace{-10pt}
    
    \label{supp:application_i2i}
\end{figure*}

In Figure~\ref{supp:application_i2i}, we present the results of AM-Adapter in image-to-image translation. The first and third rows showcase exemplar images representing a diverse range of weather conditions and times of day. The second and fourth rows illustrate the generated results, where the detailed appearance of the exemplar image is seamlessly transferred into the desired structure of the given target segmentation map. Notably, even within the same category, subtle variations exist. For instance, sunny conditions can vary between partly cloudy and completely clear skies, the sky can display different hues during sunset, and night scenes can have varying levels of brightness. Describing such nuanced differences using textual descriptions is inherently challenging. By using exemplar images to replace ambiguous textual descriptions, AM-Adapter can translate the appearance of the given image into the user-intended local exemplar appearance, with segmentation maps serving as anchors.

% (a) represents the target segmentation map with the structural condition. (b) presents a rainy exemplar image, (d) a sunny exemplar image, and (f) a night-time exemplar image. (c) shows the results reflecting the rainy condition of (b), (e) reflects the sunny condition of (d), and (g) reflects the night-time condition of (f). These results demonstrate the suitability of our method for image-to-image translation tasks, as it enables the transfer of weather conditions based on the exemplar image, without depending on ambiguous textual descriptions. 

\subsection{Appearance-Consistent Consecutive Video Frame Generation}
\label{sec:appendix_appl_video}
As illustrated in Figure~\ref{supp:application_consistent}, AM-Adapter effectively generates appearance-consistent consecutive video frames when provided with the target segmentation maps for each frame and a single exemplar image. The first row depicts the target segmentation maps of consecutive frames provided by the BDD100K~\cite{yu2020bdd100k} dataset, arranged in temporal order from left to right, while the second row shows the corresponding generated images that reflect the structure of each target map. 

It is critical to note that, as mentioned in Section \textcolor{NavyBlue}{3.5}, our method was trained using pairs of images generated by applying random augmentations, such as flipping and cropping, to a single anchor image. This training process was designed to facilitate local appearance transfer, without explicitly considering inter-frame continuity or consistency. Nevertheless, our method effectively transfers the appearance of the exemplar image, resulting in consecutive frames that maintain appearance consistency. These results indicate that the performance of our model could be further refined by fine-tuning with video-image and segmentation pairs explicitly designed to ensure temporal consistency. 

\section{Limitation and Discussion}
\label{sec:appendix_limitations}
\paragrapht{Dependency on Pretraining.}
As discussed in Section \textcolor{NavyBlue}{3.5}, our training process is divided into two stages: ControlNeXt~\cite{peng2024controlnext} for generation and AM-Adapter for matching. In the first stage, ControlNeXt learns to generate realistic images using segmentation maps as conditions. However, if ControlNeXt fails to adequately learn the generative capabilities or fully grasp the semantic information from the segmentation maps, it can negatively affect the performance of the learnable matching cost in AM-Adapter, since it is learned in a data-driven manner using segmentation-image pairs. This underscores the importance of robust pre-training for ControlNeXt.

\begin{figure*}
    \centering
    \includegraphics[width=\linewidth]{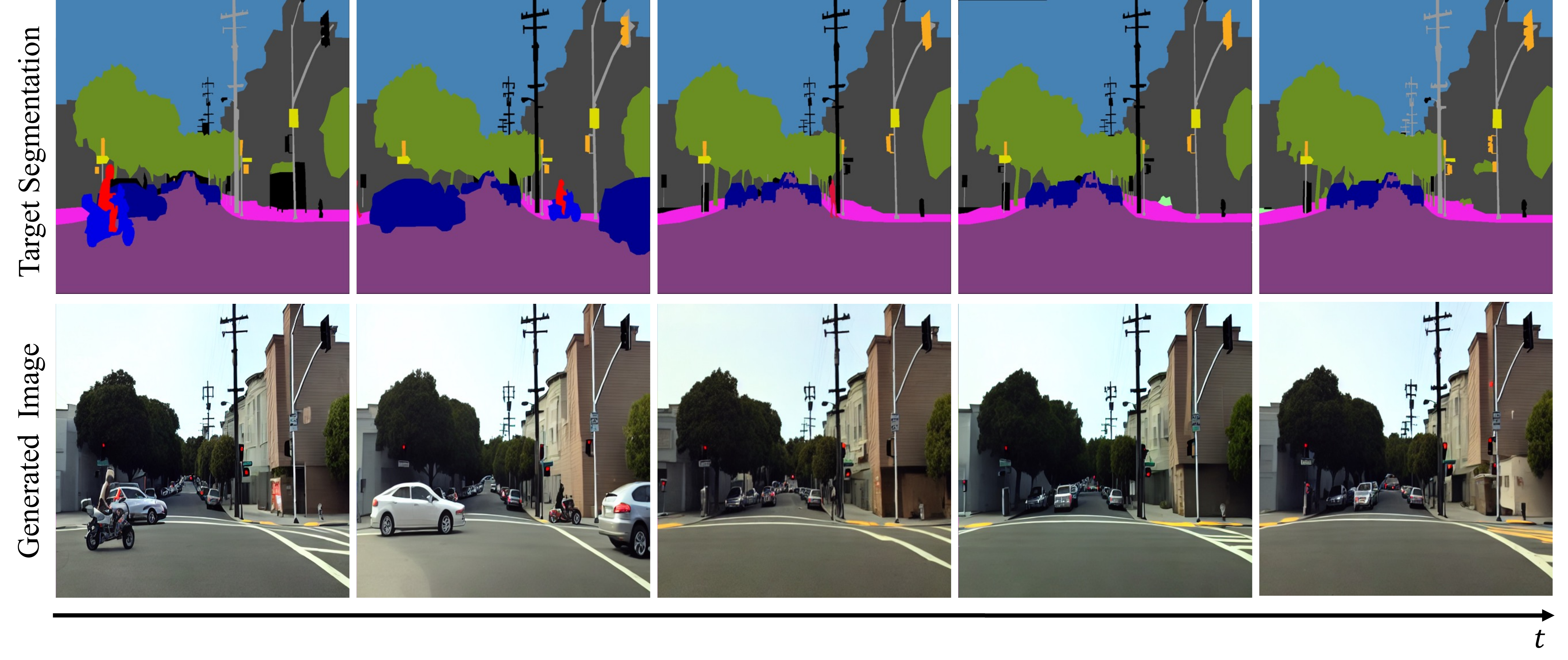} 
    \vspace{-25pt}
    \caption{\textbf{Application: Appearance-Consistent Consecutive Video Frame Generation.} The target segmentation maps in the first row are consecutive frames provided by the BDD100K~\cite{yu2020bdd100k} dataset. The second row displays the generated image results corresponding to each target segmentation maps. The frames are arranged sequentially from left to right.}
    \vspace{-10pt}
    
    \label{supp:application_consistent}
\end{figure*}

\paragrapht{Limited Temporal Consistency Under Large Scene Changes.} 
In Figure~\ref{supp:application_consistent} and Section~\ref{sec:appendix_appl_video}, we demonstrated our method’s ability to generate appearance-consistent frames using an exemplar image and the target segmentation maps of consecutive video frames as input. However, when there are substantial scene changes (e.g., large camera motion), the consistency between generated frames diminishes.

As mentioned in Section~\ref{sec:appendix_appl_video}, this limitation arises from our training approach, which uses pairs of randomly augmented images derived from a single anchor image. While this method effectively accounts for spatial differences within a pair of frames, it does not address temporal consistency across frames, resulting in temporally inconsistent images during large scene transitions. To overcome this limitation, fine-tuning the adapter with video data that explicitly incorporates temporal consistency during training could enhance its ability to generate consistent frames, even under significant scene variations.

% \section{Additional Results on SDXL}
% \label{sec:appendix_xl}
% We conducted all experiments using the SD1.5~\cite{rombach2022highresolutionimagesynthesislatent} backbone. Figure~\ref{supp:sdxl} further shows the results using SDXL~\cite{podell2023sdxlimprovinglatentdiffusion} as the backbone.

\clearpage
\newpage
% {
%     \small
%     \bibliographystyle{ieeenat_fullname}
%     \bibliography{main}
% }
\end{document}